\documentclass[runningheads]{llncs}

\usepackage{eccv}

\usepackage{eccvabbrv}

\usepackage{graphicx}
\usepackage{booktabs}
\usepackage{diagbox}
\usepackage{pifont}
\usepackage{subcaption}
\usepackage{marvosym}
\usepackage{wrapfig}
\usepackage{multirow}
\usepackage{subcaption}

\usepackage[accsupp]{axessibility}  

\usepackage{hyperref}

\usepackage{orcidlink}

\begin{document}

\title{Surface-Centric Modeling for High-Fidelity Generalizable Neural Surface Reconstruction} 

\titlerunning{Surface-Centric Modeling for High-Fidelity Surface Reconstruction}

\author{Rui Peng\inst{1,2} \and
Shihe Shen\inst{1} \and
Kaiqiang Xiong\inst{1} \and Huachen Gao\inst{1} \and \\Jianbo Jiao\inst{3} \and Xiaodong Gu\inst{4} \and Ronggang Wang\textsuperscript{\Letter}\inst{1,2}
}

\authorrunning{R. Peng et al.}

\institute{\textsuperscript{1}School of Electronic and Computer Engineering, Peking University \\ \textsuperscript{2}Peng Cheng Laboratory \quad \textsuperscript{3}University of Birmingham \quad \textsuperscript{4}Alibaba \\
\texttt{ruipeng@stu.pku.edu.cn} \quad \texttt{rgwang@pkusz.edu.cn}
}

\maketitle

\begin{abstract}
  Reconstructing the high-fidelity surface from multi-view images, especially sparse images, is a critical and practical task that has attracted widespread attention in recent years. However, existing methods are impeded by the memory constraint or the requirement of ground-truth depths and cannot recover satisfactory geometric details. To this end, we propose \textit{SuRF}, a new Surface-centric framework that incorporates a new Region sparsification based on a matching Field, achieving good trade-offs between performance, efficiency and scalability. To our knowledge, this is the first unsupervised method achieving end-to-end sparsification powered by the introduced matching field, which leverages the weight distribution to efficiently locate the boundary regions containing surface. Instead of predicting an SDF value for each voxel, we present a new region sparsification approach to sparse the volume by judging whether the voxel is inside the surface region. In this way, our model can exploit higher frequency features around the surface with less memory and computational consumption. Extensive experiments on multiple benchmarks containing complex large-scale scenes show that our reconstructions exhibit high-quality details and achieve new state-of-the-art performance, \textit{i.e.}, 46\% improvements with 80\% less memory consumption. Code is available at \url{https://github.com/prstrive/SuRF}.
\end{abstract}

\begin{figure}[t]
\centering
    \includegraphics[trim={1cm 11cm 1cm 11cm},clip,width=\textwidth]{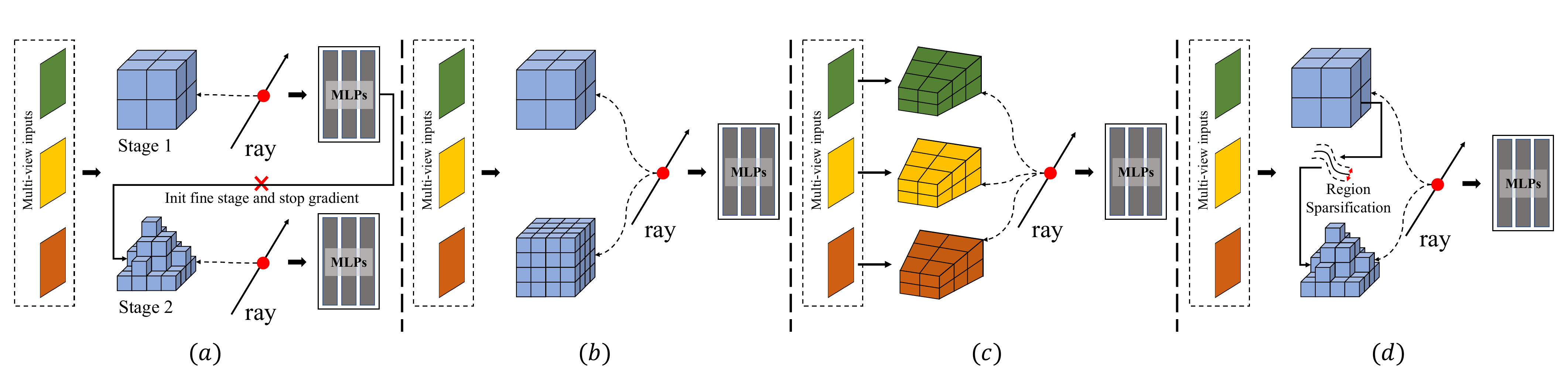}
    \captionof{figure}{{\bf Pipeline comparison with existing methods.} We omit the supervised depth-fusion methods \cite{ren2023volrecon,liang2023rethinking} and only visualize two scales or stages here for convenience. (a) The multi-stage pipeline like SparseNeuS \cite{long2022sparseneus} is not end-to-end and tends to accumulate errors, whose coarse stages cannot be optimized together with fine stages and can no longer be corrected. (b) To achieve end-to-end training, methods like GenS \cite{peng2023gens} applied the multi-scale structure to concatenate the coarse and fine volumes together, but the memory constraints limit the volume resolution. (c) View-frustum based methods like C2F2NeuS \cite{xu2023c2f2neus} construct a separate cost volume for each view, which consumes much memory and computation, especially when there are many input views. On the contrary, we design an end-to-end and sparse pipeline (d), which can leverage higher-resolution volumes with less memory and computational consumption, and the coarse model can be optimized together with the fine model.}
    \vspace{-10pt}
    \label{fig:pipe_compare}
\end{figure}

\section{Introduction}
\label{sec:intro}

Reconstructing surface from multi-view images is a fundamental and challenging task in computer vision with wide-ranging applications, including autonomous driving, robotics, virtual reality, and more. While many typical methods \cite{kazhdan2013screened,schonberger2016structure,yao2018mvsnet,gu2020cascade,peng2022rethinking} have achieved satisfactory results through tedious multi-stage processes (\ie, depth estimation, filtering and meshing), recent neural implicit methods \cite{niemeyer2020differentiable,yariv2020multiview,zhang2021learning,oechsle2021unisurf,wang2021neus,yariv2021volume} attract increasing attention due to their concise procedures and impressive reconstructions. They can directly extract the geometry through Marching Cube \cite{lorensen1998marching}, and avoid the accumulated errors. Despite their effectiveness, these methods are hampered by the cumbersome per-scene optimization and the requirement of a large number of input views, which makes them unsuitable for many applications. Even recent fast methods like \cite{wu2022voxurf,wang2023neus2} and 3D Gaussian Splatting methods \cite{kerbl20233d,chen2023neusg,guedon2023sugar} struggle to extract meshes in seconds and perform poorly under sparse input.
 
Recently, some generalizable neural surface methods \cite{long2022sparseneus,ren2023volrecon,xu2023c2f2neus,liang2023rethinking,peng2023gens} have been proposed to mitigate these problems by combining neural implicit representations with prior image information. However, as the pipeline comparisons shown in Fig. \ref{fig:pipe_compare}, they either rely on the non-end-to-end pipeline that leads to accumulated errors, or require constructing the dense volumes (or even separate volumes) for each view and consumes excessive memory and computation. We are interested in the question: {\em why unsupervised end-to-end sparsification has not been achieved yet?} To sparse the volume for the next fine model initialization, previous methods like SparseNeuS \cite{long2022sparseneus} require predicting SDF values for a large number of voxels and determining whether the SDF values are within a threshold. This is a time-consuming operation (about 10s), making it impossible to train the coarse and fine stages together.
We note that some concurrent methods \cite{hong2023lrm,li2023instant3d} directly use a large reconstruction model to achieve sparse reconstruction, but these methods are computational expensive and can only generate the low-resolution 3D representation, thus limiting their reconstruction fidelity.

In this paper, we present SuRF, the first attempt, to our knowledge, towards simultaneously unsupervised, sparsified and end-to-end approach, which provides good trade-offs between performance, efficiency, and scalability. The main idea behind this is the surface-centric modeling we adopt, which focuses more attention on regions near the surface, called ``surface regions'', a practice that improves both performance and efficiency. On the one hand, the projection feature in surface regions is more multi-view consistent and more useful for geometric reasoning. On the other hand, since the surface region only occupies a small proportion of the scene bounding box, this focusing strategy can obviously save memory and computational overhead, and enable the usage of high-resolution volumes. To achieve this, we design a module called Matching Field to locate surface regions, which poses two advantages: 1) it is the first to use the weight distribution along rays to represent the geometry, and enable the use of image warping loss to achieve unsupervised training; 2) it is highly efficient that only needs an additional single-channel volume and the very-fast trilinear interpolation. Concretely, at each scale, in addition to the n-channel feature volume used for final geometric inference, we construct another single-channel matching volume for predicting the matching field.

\begin{figure}[t]
\centering
    \includegraphics[trim={2cm 9.5cm 3.5cm 14cm},clip,width=\textwidth]{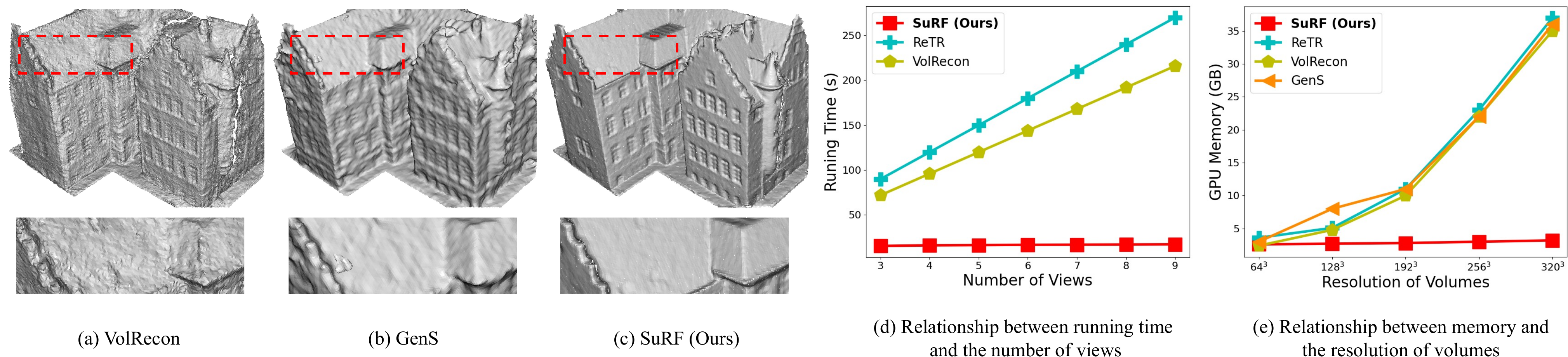}
    \captionof{figure}{{\bf Comparisons against recent state-of-the-art methods.} All experiments were conducted under the same configuration, \eg, $600\times 800$ resolution and 512 rays. The reconstruction of our method is more accurate and detailed. While the memory consumption of other methods \cite{long2022sparseneus,ren2023volrecon,liang2023rethinking,xu2023c2f2neus} increases exponentially with volume resolution, we can utilize higher-resolution feature volumes with smaller memory overhead to reconstruct higher-frequency details.
    Meanwhile, our method can directly extract meshes using Marching Cubes on the SDF like \cite{peng2023gens,long2022sparseneus}, whose consumption is more stable with vary input numbers, and is more efficient than depth-fusion methods \cite{ren2023volrecon,liang2023rethinking}.
    }
    \vspace{-10pt}
    \label{fig:head_fig}
\end{figure}

Based on the matching field, we propose a new strategy called Region Sparsification to generate sparse volumes for later high-resolution scales. Instead of predicting the SDF values for each voxel using MLPs like existing methods, we retain only voxels in surface regions visible from at least two views, which can circumvent the influence of occlusion. Thus, we can generate multi-scale and surface-centric feature volumes to remarkably improve the geometric details of the reconstruction with less memory and computational consumption, as shown in Fig. \ref{fig:head_fig}. Extensive experiments on DTU \cite{aanaes2016large} BlendedMVS \cite{yao2020blendedmvs}, Tanks and Temples \cite{knapitsch2017tanks} and ETH3D \cite{schops2017multi} datasets validate the efficiency of the proposed model, surpassing the baseline model \cite{long2022sparseneus} by more than 46\% and saving more than 80\% memory consumption compared with previous state-of-the-art methods \cite{peng2023gens,ren2023volrecon}. In summary, our main contributions are highlighted below:
\begin{itemize}
\item We make the first attempt to achieve unsupervised end-to-end sparsification in neural surface model for high-fidelity sparse reconstruction. 
\item We present a novel matching field to locate surface regions, which apply the weight distribution to represent the geometry and use image warping loss to achieve unsupervised training.
\item We introduce a new region sparsification strategy based on the extracted surface region that is robust to occlusions.
\item Extensive experiments on standard benchmarks validate the effectiveness of our approach from the perspectives of accuracy, efficiency and scalability.
\end{itemize}

\section{Related Works}
\label{sec:relatedworks}

{\bf Multi-view stereo.} Multi-view stereo (MVS) is a type of methods that take the stereo correspondence as the main cue to reconstruct geometry from multi-view images. Taking the scene representation as an axis of taxonomy, it can be broadly categorized into three types: voxel grids-based \cite{kutulakos2000theory,seitz1999photorealistic}, point clouds-based \cite{furukawa2009accurate,lhuillier2005quasi}, and depth map-based \cite{schonberger2016pixelwise,galliani2015massively,campbell2008using}.  Among them, depth map-based methods decompose complex 3D reconstructions into explicit 2D depth map estimates, becoming the most common one due to convenience. In particular, many learning-based methods \cite{yao2018mvsnet,gu2020cascade,peng2022rethinking} have been proposed to improve the matching accuracy through a more robust cost volume. However, the surface reconstruction of these methods is based on a multi-stage pipeline, which is cumbersome and inevitably introduces accumulated errors.

\noindent
{\bf Neural surface reconstruction.} Although previous volumetric methods \cite{niessner2013real} have achieved high-quality reconstructions, neural implicit functions have recently revealed significant potential in 3D reconstruction \cite{mescheder2019occupancy,park2019deepsdf,yariv2020multiview,zhang2021learning,wang2021neus,yariv2021volume,oechsle2021unisurf} and appearance modeling \cite{zhang2021nerfactor,mildenhall2021nerf,barron2021mip,oechsle2019texture,muller2022instant,wang2021ibrnet}. Some work \cite{niemeyer2020differentiable,yariv2020multiview,zhang2021learning,zhang2022critical} apply surface rendering to reconstruct plausible geometry without 3D supervision, but they often require extra priors like object masks \cite{niemeyer2020differentiable,yariv2020multiview} or sparse points \cite{zhang2022critical}. Inspired by the success of NeRF \cite{mildenhall2021nerf} in novel view synthesis, more and more methods integrate volume rendering into shape modeling. They treat the density of volume rendering as the function of different implicit representations, \eg, \cite{oechsle2021unisurf} adopts the occupancy network to represent the geometry and \cite{wang2021neus,yariv2021volume,wang2022hf,wu2022voxurf} apply the signed distance function to replace the local transparency function. Nevertheless, such methods suffer from lengthy per-scene optimization, cannot generalize to new scenes and perform poorly with sparse inputs.

\noindent
{\bf Generalizable neural surface reconstruction.} Similar to the generalizable novel view synthesis methods \cite{wang2021ibrnet,yu2021pixelnerf,chen2021mvsnerf,johari2022geonerf}, several methods \cite{long2022sparseneus,ren2023volrecon,liang2023rethinking,xu2023c2f2neus} are proposed to solve the generalization of neural surface reconstruction. By replacing the input from spatial coordinates with image features, these methods can achieve impressive cross-scene generalization. Method \cite{long2022sparseneus} is the first attempt to achieve this through a multi-stage pipeline, but still struggles to recover geometric details. Even recent methods have tried to improve this through the view-dependent representation \cite{ren2023volrecon}, transformer architecture \cite{liang2023rethinking} and even build a separate cost volume for each view \cite{xu2023c2f2neus}, still failing to balance performance, efficiency, and scalability. To be specific, these methods are restricted by the requirement of the ground-truth depth \cite{ren2023volrecon,peng2022rethinking}, cannot use high-resolution feature volumes \cite{long2022sparseneus,ren2023volrecon,peng2022rethinking,xu2023c2f2neus} and cannot scale to cases with more input views \cite{xu2023c2f2neus} due to memory constraints. In this paper, we propose SuRF, which can reconstruct more geometric details with limited memory consumption.

\section{Methodology}
\label{sec:methodology}

\begin{figure}[t]
\centering
    \includegraphics[trim={5.7cm 9.5cm 13cm 11cm},clip,width=\textwidth]{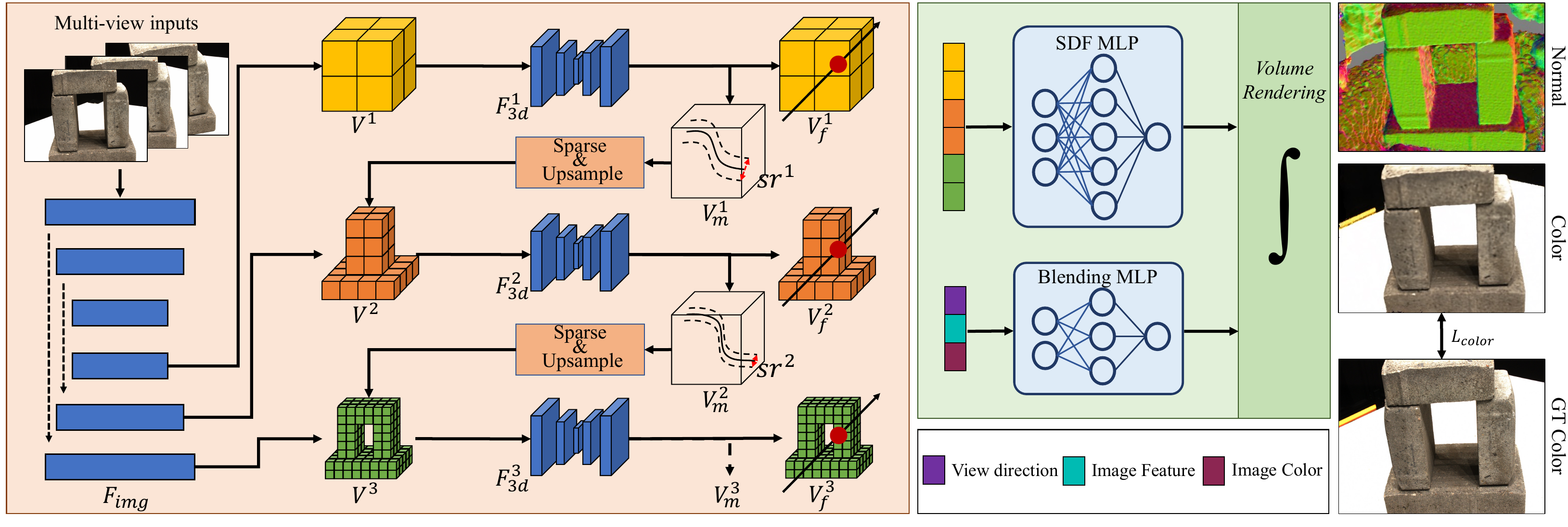}
    \caption{{\bf Framework of SuRF.} The multi-scale features are extracted through an FPN network to generate the global volume through our cross-scale fusion strategy. We then build our multi-scale surface-centric feature volumes through the region sparsification, which is based on the surface region extracted from the matching field. We employ color blending to estimate the appearance of points sampled by the surface sampling, and adopt volume rendering to recover the color of a pixel. Here, we omit some modules, \eg, surface sampling and cross-scale fusion, for convenience.}
    \vspace{-10pt}
    \label{fig:model}
\end{figure}

In this paper, our goal is to reconstruct the finely detailed and globally smooth surface $\mathcal{S}$ from an arbitrary number of inputs with limited memory and computational consumption, which is achieved through our surface-centric modeling. The overall framework of our model is illustrated in Fig. \ref{fig:model}. We first introduce our overall pipeline in Sec. \ref{sec:overview}, including how to aggregate multi-view features and reason about geometry and appearance. Then we depict our matching field in Sec. \ref{sec:matching_field}, including the unsupervised training and surface regions localization, and detail how to construct the multi-scale surface-centric feature volumes based on our new sparsification strategy in Sec. \ref{sec:coarse2fine}. The combination of the final loss function is described in Sec. \ref{sec:loss}.

\subsection{Overall Pipeline}
\label{sec:overview} 

Given a set of calibrated images $\{I_i \in \mathbb{R}^{3 \times H \times W}\}_{i=1}^N$ captured from $N$ different viewpoints, we first extract the multi-scale features $\{F_i^j \in \mathbb{R}^{C \times H \times W}\}_{i,j=1,1}^{N,L}$ through a weight-shared FPN \cite{lin2017feature} network $\mathcal{F}_{img}$. To aggregate these multi-view features, we adopt an adaptive cross-scale fusion strategy, which can grasp both global and local features and is more robust to occlusion.

\noindent
{\bf Cross-scale fusion.} For a volume $V$ with $U$ number of voxels, we project each voxel $\mathbf{v}=(x,y,z)$ to the pixel position of corresponding viewpoint with camera intrinsics $\{K_i\}_{i=1}^N$ and extrinsic $\{[R,\mathbf{t}]_i\}_{i=1}^N$:
\begin{equation}
\label{eq:w2p}
\mathbf{q}_i=\pi(K_iR_i^T(\mathbf{v}-\mathbf{t}_i)),
\end{equation}
where $\pi([x,y,z]^T)=[x/z,y/z]^T$. The corresponding multi-scale features $\{\mathbf{f}_i^j \in \mathbb{R}^C\}_{i,j=1,1}^{N,L}$ are then sampled from all image planes via bilinear interpolation. We treat the high-scale features as detail residuals of low-scale features, and sum them together as multi-view features $\{\mathbf{f}_i \in \mathbb{R}^C\}_{i=1}^N$, which are then input to a fusion network $\mathcal{F}_{fus}$ to generate view's fusion weights $\{w_i\}_{i=1}^N$. The final fused feature for each voxel is the concatenation of weighted mean and variance features $[Mean(\mathbf{v}), Var(\mathbf{v})]$:
\begin{equation}
\label{eq:fuse}
Mean(\mathbf{v})=\sum_{i=1}^Nw_i\mathbf{f}_i, \ Var(\mathbf{v})=\sum_{i=1}^Nw_i(\mathbf{f}_i-Mean(\mathbf{v}))^2.
\end{equation}

Further regularizing above fused features through a 3D network $\mathcal{F}_{3d}$, we can get the final single-channel matching volume $V_m \in \mathbb{R}^{1\times U}$ and n-channel feature volume $V_f \in \mathbb{R}^{C'\times U}$.  Note that in our surface-centric modeling, we will generate the multi-scale feature volumes $\{V_f^j\}_{j=1}^L$ and only voxels in the surface regions will be retained at high-resolution scales. The detailed procedure of surface region localization using our matching field will be stated in Sec. \ref{sec:matching_field}, and how to construct the multi-scale surface-centric feature volumes using our new region sparsification strategy will be elaborated in Sec. \ref{sec:coarse2fine}.

In this way, the surface can be reconstructed by the zero-level set of SDF values, which is estimated through a surface prediction network $\mathcal{F}_{sdf}$, which concatenate interpolations of multi-scale feature volumes as input:
\begin{equation}
\label{eq:sdf}
\mathcal{S}=\{\mathbf{p}\in \mathbb{R}^3 | \mathcal{F}_{sdf}(\mathbf{p},<\{V_f^j(\mathbf{p})\}_{j=1}^L>)=0\},
\end{equation}
where $<\cdot>$ is a concatenation operator. Meanwhile, since the traditional sampling operator cannot interpolate from the sparse volume, we implement a sparse trilinear sampling algorithm to achieve interpolation efficiently. Inherited from \cite{wang2021ibrnet}, most methods employ a similar blending strategy to predict the color of each point on a ray:
\begin{equation}
\label{eq:sdf}
\mathbf{c}=\sum_{i=1}^N\eta_i\hat{\mathbf{c}_i},
\end{equation}
where $\{\hat{\mathbf{c}_i}\}_{i=1}^N$ is the projected colors from source views, and $\{\eta\}_{i=1}^N$ is the softmax-activated blending weights estimated through a color prediction network $\mathcal{F}_{color}$,
which takes projected image features and viewing direction differences as input.

Finally, alpha-composition of samples $\{\mathbf{p}(t_k)=\mathbf{o}+t_k\mathbf{d}|k=1,...,M\}$ is performed to produce the color of a ray emitting from camera center $\mathbf{o}$ in view direction $\mathbf{d}$:
\begin{equation}
\hat{C}=\sum_{k=1}^{M} T_k\alpha_k\mathbf{c}_k,\  T_k=\prod_{l=1}^{k-1}(1-\alpha_l),
\end{equation}
where $\alpha$ is formulated in an unbiased and occlusion-aware conversion of SDF values:
\begin{equation}
\alpha_k=\max(\frac{\Phi_s(\mathcal{F}_{sdf}(\mathbf{p}(t_k)))-\Phi_s(\mathcal{F}_{sdf}(\mathbf{p}(t_{k+1})))}{\Phi_s(\mathcal{F}_{sdf}(\mathbf{p}(t_k)))},0),
\end{equation}
where $\Phi$ is the sigmoid function and $s$ is an anneal factor, please refer to \cite{wang2021neus} for more details.

\begin{figure}[t]
\centering
    \includegraphics[trim={2cm 11.5cm 24.5cm 9.5cm},clip,width=\textwidth]{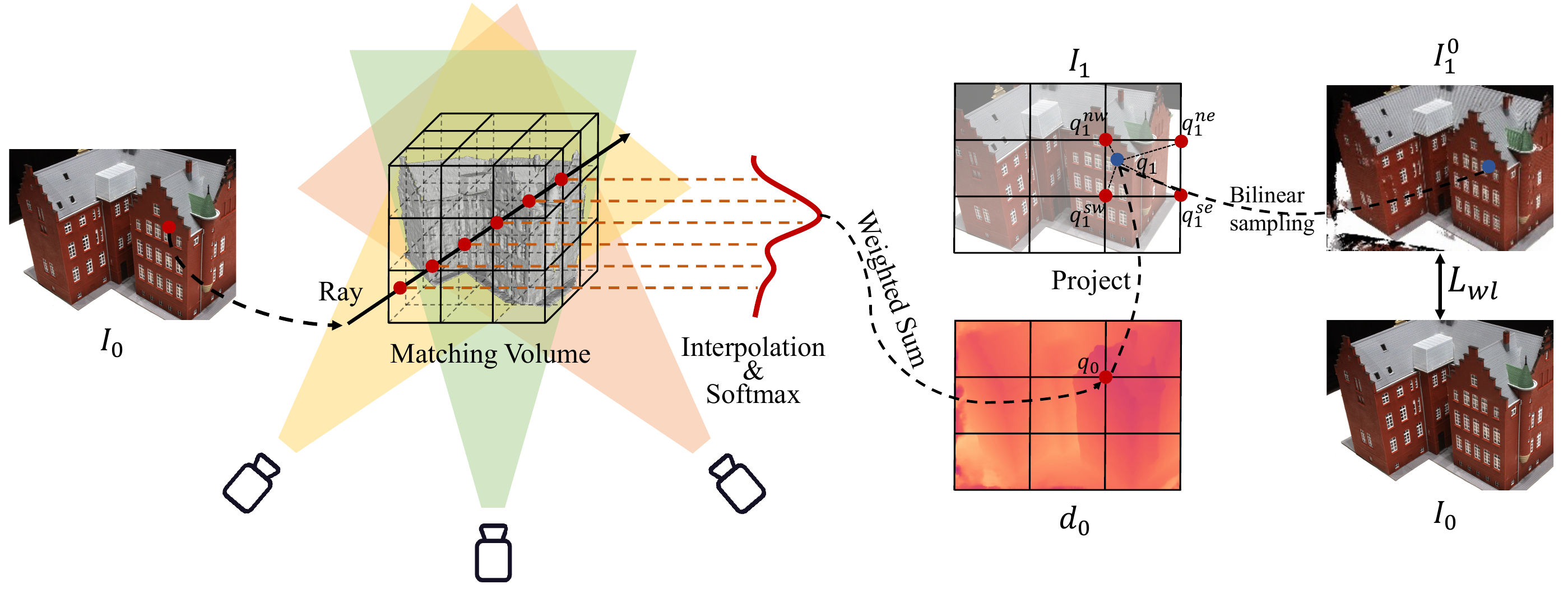}
    \vspace{-15pt}
    \caption{{\bf Illustration of our matching field.} We encode the rough scene geometry into a matching volume, and the surface position of a ray can be efficiently retrieved through interpolation. For convenience, we illustrate the surface map $E_0$ corresponding to all rays of image $I_0$ in the form of a depth map $d_0$. Then we leverage the warping loss $L_{wl}$ to constrain the matching field unsupervised.}
    \vspace{-10pt}
    \label{fig:match_field}
\end{figure}

\subsection{Matching Field}
\label{sec:matching_field}

Fig. \ref{fig:match_field} illustrates the overall pipeline of our matching field, which is the cornerstone of surface-centric modeling. In this section, we will elaborate on it in twofold: how it achieves the surface region localization, and how it achieves unsupervised training. Note that these procedures are the same for all scales, and we omit the subscript of scales for convenience.

\noindent
{\bf Surface region localization.} For efficiency, we need this procedure to hold three important properties: 
\begin{itemize}
\item It requires encoding the entire scene geometry with limited memory consumption
. Existing methods like \cite{johari2022geonerf} or \cite{xu2023c2f2neus} borrow the main idea of MVS methods \cite{gu2020cascade} to construct a separate cost volume for each view, which is sometimes impractical for surface reconstruction, especially when there are many input views.
\item It needs to rapidly locate the surface region with a small computational cost, which makes multi-stage training or the use of extra networks unworkable.
\item It needs to be occlusion-aware and view-dependent, \ie, those surfaces that are behind or not visible from the input views are unuseful and unsolvable.
\end{itemize}

Motivated by these properties, we implement the matching field as a weight distribution along the ray obtained from matching volume interpolation.

As shown in Fig. \ref{fig:match_field}, instead of representing the geometry as the occupancy, density or SDF value, we employ the view-dependent weight distribution, where larger values represent closer proximity to the surface. Concretely, to extract the surface of a ray $\mathbf{r}=(\mathbf{o},\mathbf{d})$, we first uniformly sample $M_s$ points $\{\mathbf{p}(t_k)=\mathbf{o}+t_k\mathbf{d}\}_{k=1}^{M_s}$ within the current surface region (Note that $M_s$ decreases as the scale increases). Next, we directly interpolate the corresponding value for each point from the matching volume $V_m$, and then go through a softmax operator to generate the weight distribution $\{\gamma_k\}_{k=1}^{M_s}$ along this ray. In this way, we can infer the rough position of the surface point $\mathbf{p}_s=\mathbf{o}+t_s\mathbf{d}$, where:
\begin{equation}
t_s=\sum_{k=1}^{M_s}\gamma_kt_k.
\end{equation}

Finally, the surface region that we need is defined as: $sr=[t_s-\epsilon, t_s+\epsilon]$, and $\epsilon$ is a hyperparameter that gradually decreases as the scale increases. And the surface region is set to the length of scene bounds for the first scale.

\noindent
{\bf Unsupervised training.} With the surface points, we can conveniently leverage the image warping loss \cite{godard2019digging,peng2021excavating} to constrain the matching field. Supposing the reference image $I_0$ has a resolution of $H\times W$, through the matching field, we can efficiently retrieve the surface point of all rays emitting through the pixels of $I_0$ to form the ``surface map'' $E_0\in \mathbb{R}^{3\times H\times W}$. Then we project these points to the pixel positions of source images $\{I_i\}_{i=1}^N$ through Eq. (\ref{eq:w2p}), and interpolate the colors to generate the warped images $\{I_i^0\}_{i=1}^N$. Theoretically, the projected colors of these surface points should remain consistent across multiple viewpoints. Therefore, we can generate the constraints through the difference between the ground-truth reference image and the warped images from source images. Furthermore, we combine the pixel-wise color loss with the patch-based SSIM \cite{wang2004image}:
\begin{equation}
WL_i=0.8 \times \frac{1 - SSIM(I_0, I_i^0)}{2}+0.2 \times |I_0-I_i^0|.
\end{equation}

To avoid the influence of occlusions, we take the average of the K smallest warping losses as the final constraint to optimize our matching field:
\begin{equation}
L_{wl}=\frac{1}{K}\sum_{i=1}^KWL_i.
\end{equation}

In this way, we can optimize our matching field unsupervised to locate the surface region efficiently.

\begin{figure}[t]
    \centering
    \includegraphics[trim={1cm 11cm 4cm 11cm},clip,width=1.0\linewidth]{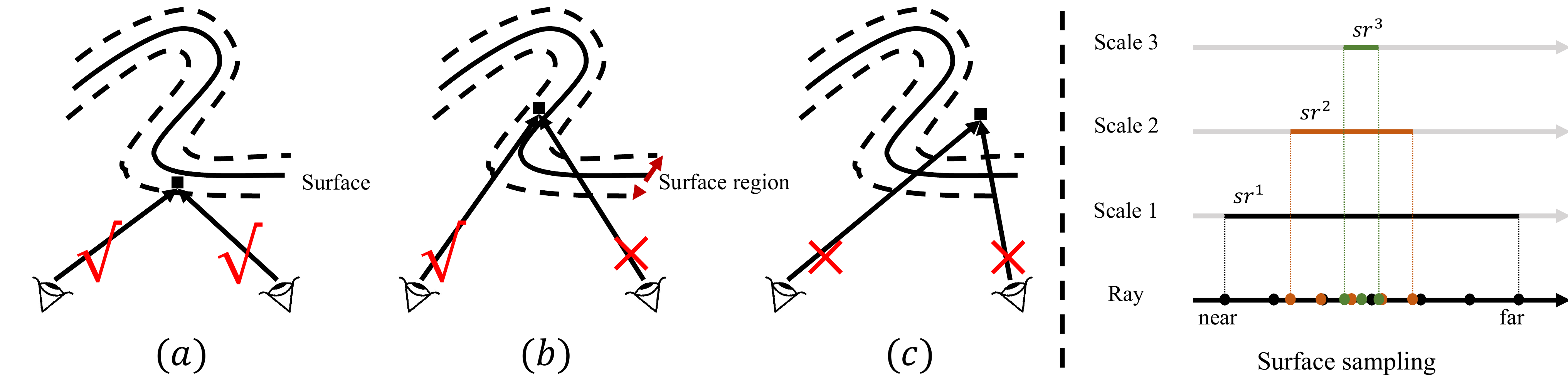}
    \vspace{-10px}
    \caption{{\bf Illustration of region sparsification and surface sampling.} For sparsification, we illustrate three situations: voxels that fall into surface regions visible from multiple viewpoints (a) will be preserved; voxels that fall into surface regions only visible by one viewpoint (b) or are outside surface regions (c) will be pruned. The black rectangle represents the position of a voxel. For surface sampling, we sample the points for each ray within the surface region at each scale.}
    \vspace{-10pt}
    \label{fig:filtering}
\end{figure}

\subsection{Feature Volume Construction based on Region Sparsification}
\label{sec:coarse2fine}

As aforementioned and the pipeline comparisons shown in Fig. \ref{fig:pipe_compare}, previous methods \cite{peng2023gens,xu2023c2f2neus} rely on the dense volume or multi-stage training \cite{long2022sparseneus}, and they either are limited by the memory constraints or introduce cumulative errors. To this end, based on the surface region located through our matching field, we propose a region sparsification strategy to construct the multi-scale and surface-centric feature volumes to mitigate these drawbacks.

\noindent
{\bf Region sparsification.} Taking a certain scale $j$ as an example, assume that we have generated the matching volume $V_m^j\in \mathbb{R}^{1\times U^j}$ and feature volume $V_f^j\in \mathbb{R}^{C'\times U^j}$ according to the process in Sec. \ref{sec:overview}, and obtained the surface maps $\{E_i^j\in \mathbb{R}^{3\times H\times W}\}_{i=1}^N$ of all views following the pipeline in Sec. \ref{sec:matching_field}. To prune those voxels away from the surface, we project all voxels ${Vox}^j=\{\mathbf{v}_h\}_{h=1}^{U^j}$ to the pixel position of all surface maps through Eq. (\ref{eq:w2p}) and interpolate the corresponding surface points $\{\hat{E}_i^j \in \mathbb{R}^{3\times U^j}\}_{i=1}^N$ visible from each view through bilinear sampling. We then can determine whether the voxel is inside the surface region based on the distance between the voxel and the interpolated surface point:
\begin{equation}
H_i^j(\mathbf{v})=\mathtt{float}(\parallel \hat{E}_i^j(\mathbf{v})-\mathbf{v}\parallel_2 < \epsilon^j),
\end{equation}
where $\mathtt{float}$ is the operator that converts bool values to float values. Furthermore, to maintain the view consistency, we only retain voxels that simultaneously fall into the surface region of at least two views:
\begin{equation}
{Vox}^{j+1}=\{\mathbf{v} | \mathtt{sum}(H^j(\mathbf{v}))\ge 2\},
\end{equation}
where $\mathtt{sum}$ is the summation operator. This is an important step to mitigate the impact of occlusion, as regions visible only from a small number of views are meaningless, and we depict some examples in Fig. \ref{fig:filtering}. Then we can halve these surviving voxels to aggregate higher-frequency information for the next scale.

Repeating the region sparsification for each scale, we can generate the final multi-scale feature volumes $\{V_f^j\}_{j=1}^L$. While this multi-scale strategy is beneficial for the model to reconstruct surfaces with high-frequency detail and global smoothness like \cite{yu2022monosdf,peng2023gens}, our volumes are surface-centric and can achieve higher resolution with less memory consumption. Meanwhile, since the surface region in the coarse stage is wide, using multi-scale features to predict the geometry makes the model more robust when the surface region location in the fine stage is wrong. Before employing the volume rendering to produce the color of a ray, we propose surface sampling to efficiently sample more points for surface regions. 

\noindent
{\bf Surface sampling.} With the off-the-shelf surface regions $\{sr^j\}_{j=1}^L$ provided by the matching field, it's natural to sample more points within these regions because voxels inside these regions contain the most valuable information about the surface. We uniformly sample a decreasing number of points within the surface region from low-resolution to high-resolution scales, which results in more sampling points near the surface as shown in Fig. \ref{fig:filtering}. When interpolating from multi-scale feature volumes, we fill the feature of those sampling points that are outside the surface region of certain scales with zero. Therefore, we do not need other networks to resample more fine points like \cite{wang2021neus,long2022sparseneus,mildenhall2021nerf}.

\subsection{Loss Function}
\label{sec:loss}

Our overall loss function consists of three components:
\begin{equation}
L=L_{surf} + L_{mf},
\end{equation}
where $L_{surf}$ is used to optimize surface network and $L_{mf}$ is used to optimize multi-stage matching fields. 

Following existing methods \cite{long2022sparseneus,xu2023c2f2neus}, our surface loss $L_{surf}$ is defined as:
\begin{equation}
L_{surf}=L_{color} + L_{mfc} +\alpha L_{ek} + \beta L_{pe},
\end{equation}
where $L_{color}$ is computed as the average color loss of all sampled pixels $Q$:
\begin{equation}
L_{color}=\frac{1}{|Q|}\sum_{q\in Q} |C(q)-\hat{C}(q)|,
\end{equation}
$L_{mfc}$ is the feature consistency following \cite{peng2023gens}, and eikonal loss \cite{icml2020_2086} is:
\begin{equation}
L_{ek}=\frac{1}{|P|}\sum_{p\in P}(||\nabla \mathcal{F}_{sdf}(p)||_2-1)^2,
\end{equation}
where $P$ is a set of sampled 3D points. Similar to \cite{xu2023c2f2neus}, we leverage the pseudo label generated from the unsupervised multi-view stereo method \cite{chang2022rc} to enhance and accelerate model convergence. We apply a very strict filtering strategy to obtain relatively accurate pseudo point clouds $\hat{P}$. The pseudo loss is:
\begin{equation}
L_{pe}=\frac{1}{|\hat{P}|}\sum_{p\in \hat{P}}|\mathcal{F}_{sdf}(p)|.
\end{equation}

The loss of the matching field is defined as the weighted sum of all scales' warping loss:
\begin{equation}
L_{mf}=\sum_{j=1}^L\mu^j L_{wl}^j,
\end{equation}
where $\mu_j$ is the weight of stage $j$, and it increases from coarse to fine scale.

\begin{figure*}[t]
\centering
    \includegraphics[trim={4.5cm 11cm 7.5cm 1.4cm},clip,width=\textwidth]{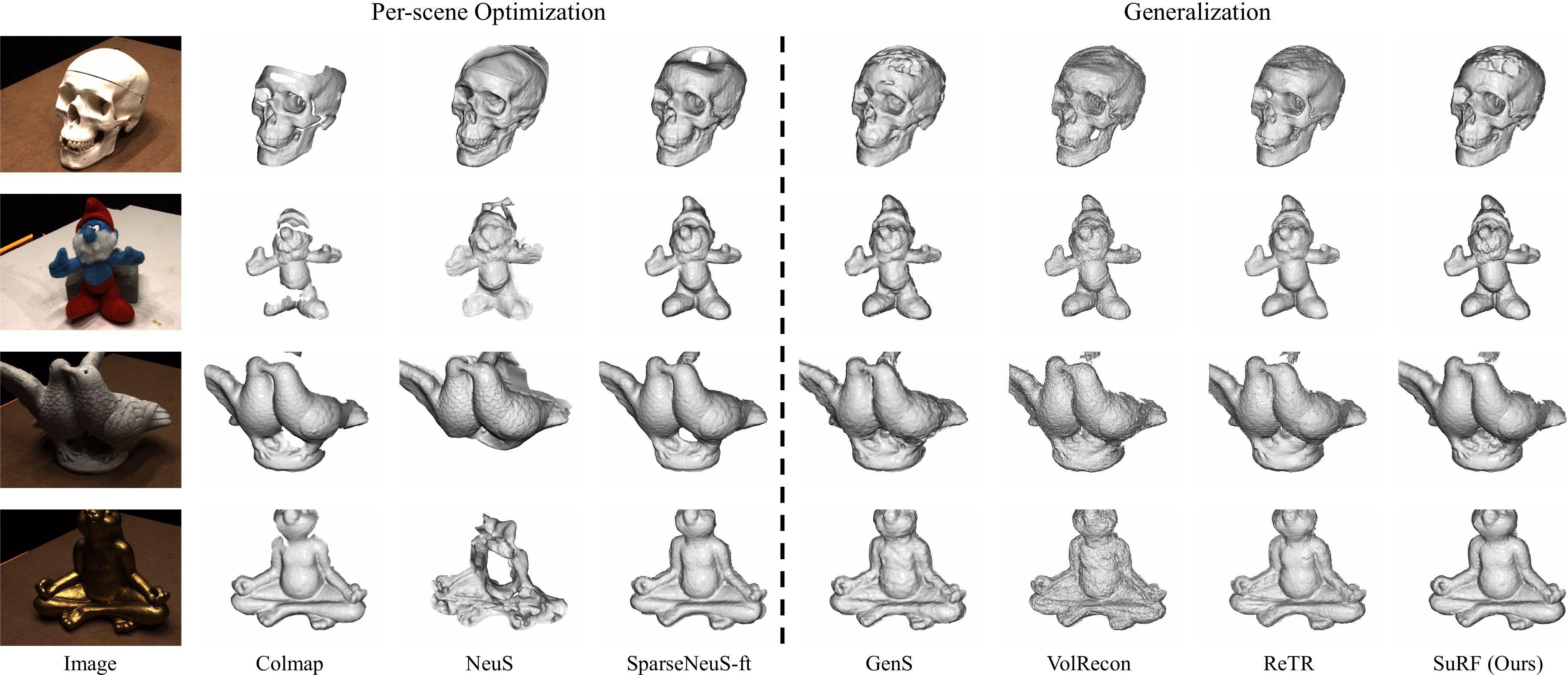}
    \vspace{-15pt}
    \caption{{\bf Qualitative comparisons on DTU dataset.}}
    \vspace{-15pt}
    \label{fig:dtu_res}
\end{figure*}

\section{Experiments}
\label{sec:experiments}

In this section, we first introduce our implementation details and compared datasets, then we enumerate extensive experiments and ablation studies. Note that the reported results here are based on our model without any finetuning. Please refer to Supp. Mat. for the finetuning results.

\noindent
{\bf Implementation details.}
We implement our model in PyTorch \cite{paszke2019pytorch}, and build our surface-centric feature volume in $L=4$ scales. The range of surface regions for each scale is defined as $\epsilon^1$:$\epsilon^2$:$\epsilon^3$:$\epsilon^4$=1:0.3:0.1:0.01. In our matching field, we set the number of sampling points for each scale as 128, 64, 32, 16 to extract surface regions. To apply the final volume rendering, the total number of sampling points of each ray is set to $M=120$, which consists of 64, 32, 16 and 8 for our surface sampling. During training, we adopt Adam optimizer \cite{kingma2014adam} to train our model for 10 epochs. The base learning rate is set to 1e-3 for feature networks and 5e-4 for MLPs. We set the number of source images as $N=4$ and resize the resolution to $640\times 480$. The volume resolution of the first low-resolution scale is set to $R^1=64\times64\times64$. The weight of each loss term is set to $\alpha=0.1$, $\beta=1.0$, and the warping loss weights of each scale are set to 0.25, 0.5, 0.75 and 1.0.  During testing, we take $N=2$ source images with a resolution of $800 \times 576$ as input, and set $R^1=80\times80\times80$. Our meshes are extracted using Marching Cubes \cite{lorensen1998marching}.

\noindent
{\bf Datasets.}
Following existing practices \cite{long2022sparseneus,ren2023volrecon}, we train our model on the DTU dataset \cite{aanaes2016large}, and we employ the same splitting strategies as \cite{long2022sparseneus}, \ie, 75 scenes for training and two sets of images from 15 non-overlapping scenes for testing. To validate our generalization ability, we further conduct some qualitative comparisons on BlendedMVS \cite{yao2020blendedmvs}, Tanks and Temples \cite{knapitsch2017tanks} and ETH3D \cite{schops2017multi} datasets.

\subsection{Results on DTU}
\label{sec:dtu_res}

\begin{figure*}[t]
\centering
    \includegraphics[trim={11.5cm 15cm 14.5cm 14cm},clip,width=\textwidth]{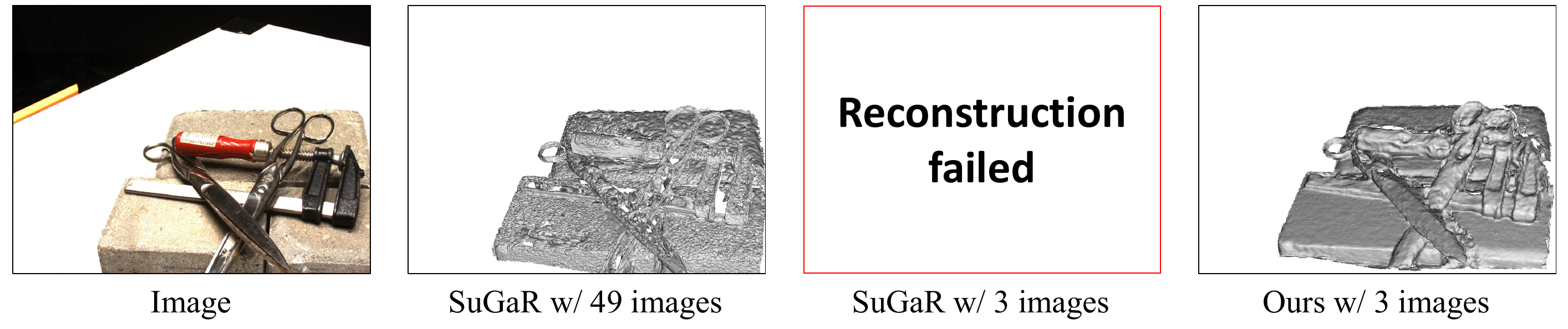}
    \vspace{-20pt}
    \caption{{\bf Qualitative comparison on DTU dataset with SuGaR.}}
    \vspace{-15pt}
    \label{fig:comp_sugar}
\end{figure*}

For a fair comparison, we adopt the same evaluation strategy as previous methods \cite{long2022sparseneus,ren2023volrecon}, \ie, reconstruct the surface using only three input views and report the average chamfer distance of two image sets.
The quantitative results on the DTU dataset are summarized in Tab. \ref{tab:dtu_res}, which indicate that our SuRF can bring a satisfactory improvement to the baseline, \ie, more than 46\% improvement compared with SparseNeuS \cite{long2022sparseneus}. Meanwhile, we can surpass those methods \cite{ren2023volrecon,liang2023rethinking} that employ ground-truth depth for supervision. Even compared with the method that construct separate cost volumes for each input view \cite{xu2023c2f2neus}, our model still offers plausible advantages in efficiency as well as scalability, that is, we only need to construct a global volume and is computation and memory insensitive to the number of input views. For those classical MVS methods \cite{schonberger2016structure,chang2022rc}, our method can win out in most metrics. Compared with the recent fast method \cite{wu2022voxurf} that converges in minutes, our model can still reconstruct finer details in seconds. Qualitative results in Fig. \ref{fig:dtu_res} further show that our SuRF can reconstruct finer surfaces only through fast network inference. The results in Fig. \ref{fig:comp_sugar} indicate that the 3DGS-based method SuGaR \cite{guedon2023sugar} failed with sparse inputs and our approach shows much better performance even with sparse inputs.

\begin{table}[t]
  \caption{{\bf Quantitative results on DTU dataset.} Best results in each category are in {\bf bold} and the second best are in \underline{underline}. `*' denotes the depth-fusion methods that need the ground-truth depth for supervision, and we report the reproduced results using their officially released model. We test \cite{chang2022rc} at the same input resolution as ours.
  }
  \label{tab:dtu_res}
  \centering
  \resizebox{1.0\linewidth}{!}{
  \begin{tabular}{l|ccccccccccccccc|c}
  \toprule
  Method & 24 & 37 & 40 & 55 & 63 & 65 & 69 & 83 & 97 & 105 & 106 & 110 & 114 & 118 & 122 & Mean \\
  \midrule
  COLMAP \cite{schonberger2016pixelwise} & \underline{0.90} & 2.89 & 1.63 & 1.08 & 2.18 & 1.94 & 1.61 & 1.30 & 2.34 & 1.28 & 1.10 & 1.42 & 0.76 & 1.17 & 1.14 & 1.52 \\
  RC-MVSNet \cite{chang2022rc} & 0.93 & 2.64 & 1.92 & 1.00 & 1.55 & 1.62 & 0.88 & 1.29 & \bf{1.16} & 1.00 & \underline{0.82} & 0.67 & 0.60 & 1.04 & 1.22 & 1.22 \\
  \midrule
  NeuS \cite{wang2021neus} & 4.57 & 4.49 & 3.97 & 4.32 & 4.63 & 1.95 & 4.68 & 3.83 & 4.15 & 2.50 & 1.52 & 6.47 & 1.26 & 5.57 & 6.11 & 4.00 \\
  VolSDF \cite{yariv2021volume} & 4.03 & 4.21 & 6.12 & 0.91 & 8.24 & 1.73 & 2.74 & 1.82 & 5.14 & 3.09 & 2.08 & 4.81 & 0.60 & 3.51 & 2.18 & 3.41 \\
  Voxurf \cite{wu2022voxurf} & 2.51 & 4.32 & 2.88 & 2.17 & 5.43 & 2.01 & 2.88 & 2.11 & 1.94 & 1.60 & 3.02 & 3.65 & 1.20 & 2.10 & 2.08 & 2.65 \\
  SparseNeuS-ft \cite{long2022sparseneus} & 1.29 & \bf{2.27} & 1.57 & \underline{0.88} & 1.61 & 1.86 & 1.06 & 1.27 & 1.42 & 1.07 & 0.99 & 0.87 & 0.54 & 1.15 & 1.18 & 1.27 \\
  \midrule
  IBRNet \cite{wang2021ibrnet} & 2.29 & 3.70 & 2.66 & 1.83 & 3.02 & 2.83 & 1.77 & 2.28 & 2.73 & 1.96 & 1.87 & 2.13 & 1.58 & 2.05 & 2.09 & 2.32 \\
  MVSNerf \cite{chen2021mvsnerf} & 1.96 & 3.27 & 2.54 & 1.93 & 2.57 & 2.71 & 1.82 & 1.72 & 2.29 & 1.75 & 1.72 & 1.47 & 1.29 & 2.09 & 2.26 & 2.09 \\
  SparseNeuS \cite{long2022sparseneus} & 2.17 & 3.29 & 2.74 & 1.67 & 2.69 & 2.42 & 1.58 & 1.86 & 1.94 & 1.35 & 1.50 & 1.45 & 0.98 & 1.86 & 1.87 & 1.96 \\
  VolRecon* \cite{ren2023volrecon} & 1.54 & 3.05 & 1.84 & 1.08 & 1.67 & 1.84 & 1.13 & 1.63 & 1.49 & 1.19 & 1.06 & 1.42 & 0.76 & 1.22 & 1.29 & 1.48 \\
  GenS \cite{peng2023gens} & 1.45 & 2.77 & 1.69 & 0.97 & 1.54 & 1.90 & 1.03 & 1.49 & 1.36 & 0.97 & 1.07 & 0.97 & 0.62 & 1.14 & 1.16 & 1.34 \\
  ReTR* \cite{liang2023rethinking} & 1.23 & 2.63 & 1.62 & 0.98 & \underline{1.38} & \underline{1.56} & 0.90 & 1.39 & 1.39 & 1.02 & 0.93 & 0.83 & 0.59 & 1.10 & 1.16 & 1.25 \\
  C2F2NeuS \cite{xu2023c2f2neus} & 1.12 & 2.42 & {\bf 1.40} & {\bf 0.75} & 1.41 & 1.77 & \underline{0.85} & \underline{1.16} & 1.26 & {\bf 0.76} & 0.91 & \underline{0.60} & {\bf 0.46} & \underline{0.88} & {\bf 0.92} & \underline{1.11} \\
  {\bf SuRF (Ours)} & {\bf 0.85} & \underline{2.35} & \underline{1.48} & \underline{0.88} & {\bf 1.17} & {\bf 1.39} & {\bf 0.74} & \bf{1.14} & \underline{1.24} & \underline{0.84} & {\bf 0.77} & \bf{0.51} & \underline{0.51} & {\bf 0.86} & \underline{1.03} & {\bf 1.05} \\

  \bottomrule
  \end{tabular}
  }
\end{table}

\noindent
{\bf Number of input views.} 
Fig. \ref{fig:num_view} indicates that the reconstruction quality of our model gradually improves as the number of views increases, but it tends to stabilize when the input views are enough. Meanwhile, a larger number of inputs does not lead to a significant increase in consumption, which is an obvious advantage compared to C2F2NeuS \cite{xu2023c2f2neus}, as the comparison shown in Tab. \ref{tab:mem_view}.

\begin{table}[t]
\centering
\begin{minipage}[c]{0.41\linewidth}
\makeatletter\def\@captype{figure}
\caption{{\bf The performance of mean chamfer distance w.r.t. different number of views.}}
\vspace{-3mm}
\label{fig:num_view}
\centering
\resizebox{1.0\linewidth}{!}{
\includegraphics[trim={1.5cm 0cm 3.1cm 1.7cm},clip,width=\textwidth]{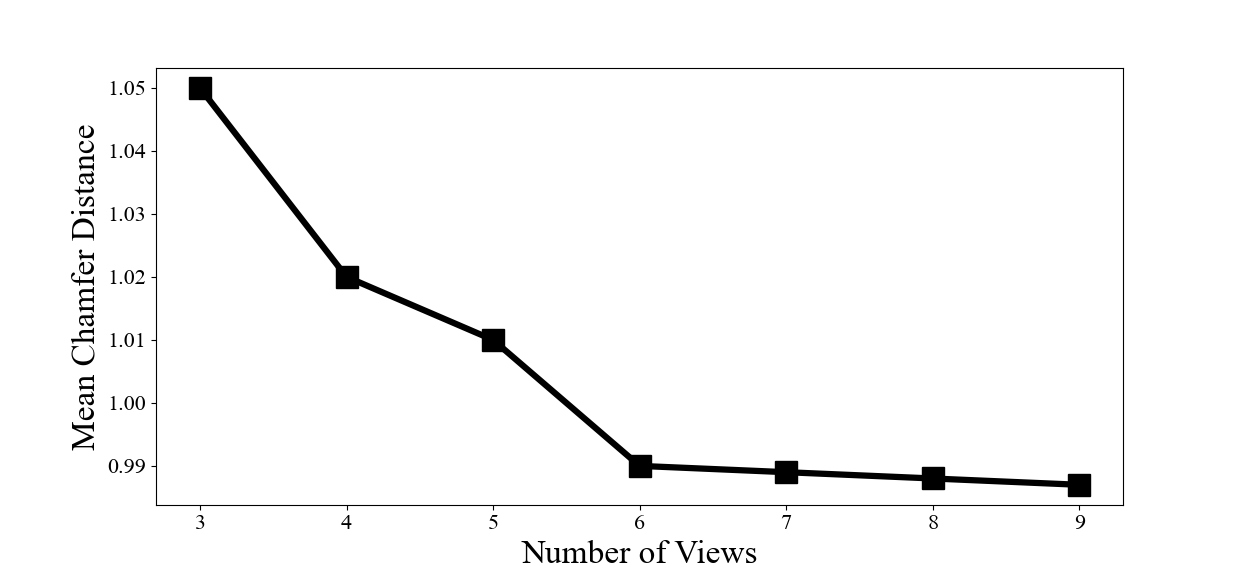}
}
\end{minipage}
\hfill
\begin{minipage}[c]{0.56\linewidth}
\makeatletter\def\@captype{table}
\caption{{\bf Memory consumption with different number of views.} Image resolution is $800\times 600$, highest volume resolution of SuRF is $256^3$.}
\label{tab:mem_view}
\centering
\resizebox{1.0\linewidth}{!}{
\begin{tabular}{c|c|c|c|c}
\toprule
\diagbox[width=14em,trim=l]{Method}{Number of views} & 3 & 5 & 7 & 9 \\
\hline
C2F2NeuS \cite{xu2023c2f2neus} & 10G & 16G & 22G & 28G \\
\hline
SuRF (Ours) & 2.6G & 2.7G & 2.8G & 3.1G \\
\bottomrule
\end{tabular}
}
\end{minipage}
\end{table}

\subsection{Generalization}
\label{sec:bmvs_res}

\begin{figure*}[t]
\centering
    \includegraphics[trim={4.5cm 11cm 15.2cm 1.4cm},clip,width=\textwidth]{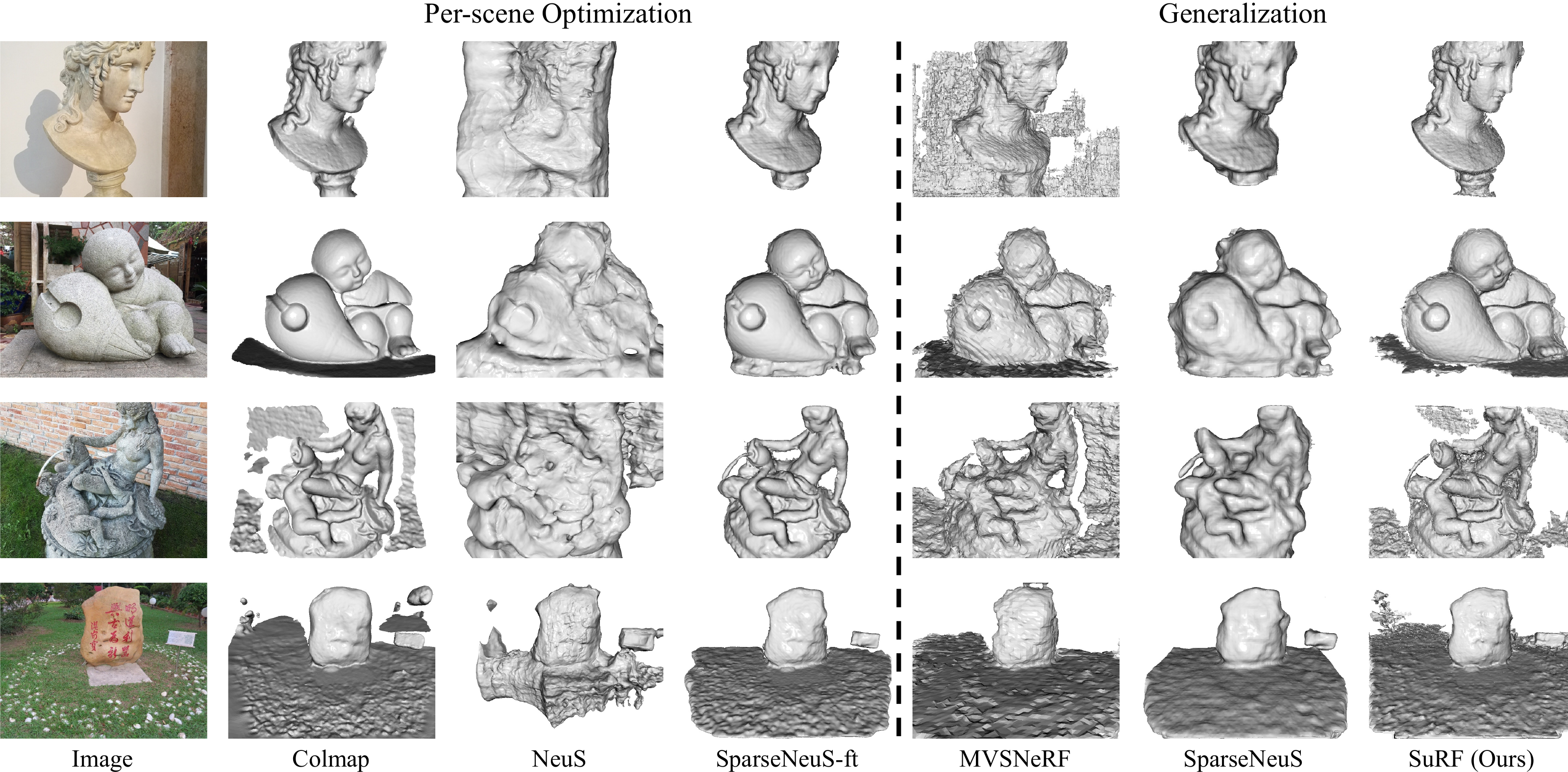}
    \vspace{-15pt}
    \caption{{\bf Qualitative comparisons on BlendedMVS dataset.}}
    \vspace{-15pt}
    \label{fig:bmvs_res}
\end{figure*}

\begin{figure*}[t]
\centering
    \includegraphics[trim={4cm 8cm 7cm 8.9cm},clip,width=\textwidth]{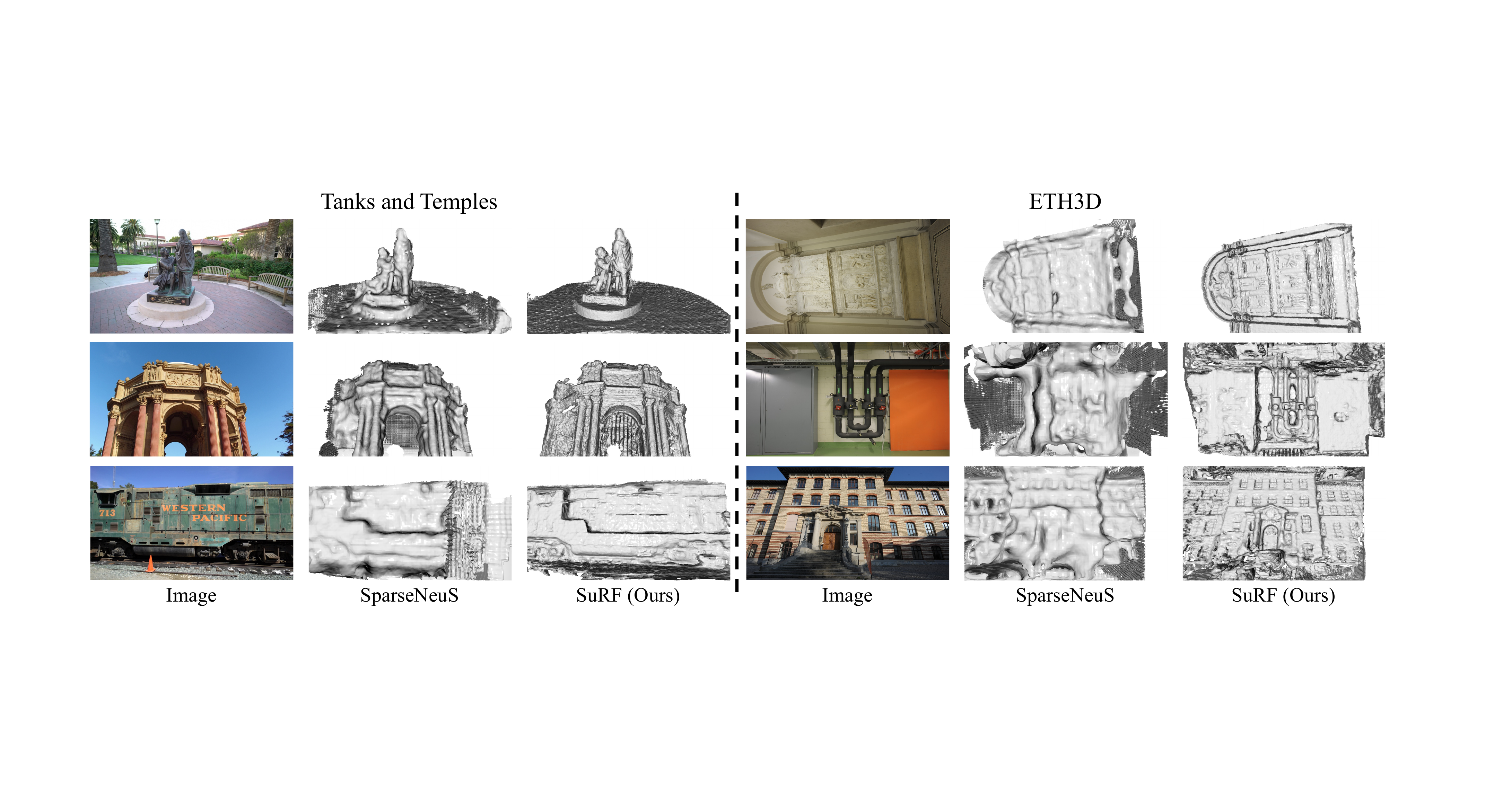}
    \vspace{-20pt}
    \caption{{\bf Qualitative comparisons on Tanks and Temples and ETH3D datasets.}}
    \vspace{-15pt}
    \label{fig:tank_eth3d_res}
\end{figure*}

To verify the generalization capabilities of our method, we further test on BlendedMVS \cite{yao2020blendedmvs}, Tanks and Temples \cite{knapitsch2017tanks} and ETH3D \cite{schops2017multi} datasets using the model pre-trained on the DTU dataset \cite{aanaes2016large}. Some qualitative comparisons with existing methods are shown in Fig. \ref{fig:bmvs_res} and Fig. \ref{fig:tank_eth3d_res}. The results prove that our method exhibits great generalization ability even under these difficult scenes. The volume-based method struggles on large scenes since most voxels are empty, but ours can still reconstruct meshes with fine details even without any finetuning.

\begin{table}[t]
    \caption{{\bf Ablation results on DTU dataset.}} 
    \vspace{-6mm}
    \label{tab:ablation}
    \centering
    \begin{subtable}{.505\linewidth}
        \centering
        \caption{The effectiveness of our end-to-end sparsification using our main contributions.}
        \vspace{-3mm}
        \resizebox{1.0\linewidth}{!}{
        \begin{tabular}{l|c}
            \toprule
            Method & Mean Cham. Dist. $\downarrow$ \\
            \hline
            baseline & 1.70 \\
            w/ multi-stage training & 1.52 \\
            w/ multi-scale dense volume & 1.31 \\
            w/ end-to-end sparse. & {\bf 1.02} \\
            \bottomrule
        \end{tabular}
        }
    \end{subtable}%
    \hfill
    \begin{subtable}{.49\linewidth}
        \centering
        \caption{The performance of the model with different resolution of volumes and input images.}
        \vspace{-3mm}
        \resizebox{1.0\linewidth}{!}{
        \begin{tabular}{c|c|c}
            \toprule
            Res. of Vol. & Res. of Image & Mean Cham. Dist. $\downarrow$\\
            \hline
            $48\times48\times48$ & $640\times480$ & 1.12  \\
            $64\times64\times64$ & $640\times480$ & 1.07  \\
            $64\times64\times64$ & $800\times576$ & 1.04  \\
            $80\times80\times80$ & $800\times576$ & {\bf 1.02}  \\
            \bottomrule
        \end{tabular}
        }
    \end{subtable} 
    \vspace{-2mm}
\end{table}

\subsection{Ablation Studies}
\label{sec:ablation}
Our ablation experiments are performed on the first image set of the DTU dataset, which is the same as \cite{long2022sparseneus}. 
To investigate the effectiveness of our end-to-end sparsification, we compare it with existing solutions, \ie, the multi-stage training strategy in SparseNeuS \cite{long2022sparseneus} and the multi-scale dense structure in GenS \cite{peng2023gens}. The results in (a) of Tab. \ref{tab:ablation} show that all these solutions bring improvements to the baseline model, and our solution performs the best mainly because our features are surface-centric and can be continuously optimized. We perform another study to understand how the resolution of volumes and images affects the model. The results in (b) of Tab. \ref{tab:ablation} show that the higher resolution favors the model but stabilizes when a certain resolution is reached. 

\noindent
{\bf Efficiency and scalability.} Compared with existing methods, our SuRF exhibits advantages in terms of efficiency and scalability as shown in Fig. \ref{fig:head_fig}. Our model can leverage volumes of higher resolution with less memory and computational consumption, \eg, only 3G memory for the model with the highest $256^3$ resolution volumes while at least 20G for previous methods \cite{long2022sparseneus,ren2023volrecon,liang2023rethinking,xu2023c2f2neus}. Compared with methods that require predicting depth maps \cite{ren2023volrecon,liang2023rethinking,yao2018mvsnet} or constructing cost volumes \cite{xu2023c2f2neus} for each view, the running time and memory consumption of our model is relatively insensitive to the number of input views as shown in Tab. \ref{tab:mem_view}, making it scalable to handle different input numbers. 
Meanwhile, the capability of using high-resolution volumes gives our method the potential to reconstruct very large-scale scenes as the results shown in Fig. \ref{fig:tank_eth3d_res}.

\section{Conclusion}
\label{sec:conclusion}
\vspace{-1mm}

In this paper, we proposed a new generalizable neural surface model, \textit{SuRF}, to accomplish high-fidelity reconstruction even from sparse inputs with satisfactory trade-offs between performance, efficiency and scalability. To the best of our knowledge, it is the first unsupervised method to achieve end-to-end sparsification based on our surface-centric modeling, which consists of a novel matching field module and a new region sparsification strategy. The proposed matching field adopts the weight distribution to represent geometry and introduces the image warping loss to achieve unsupervised training, which can efficiently locate the surface region. Then we adopted the region sparsification strategy to prune voxels outside the surface regions and generated the multi-scale surface-centric feature volumes. Extensive experiments on multiple public benchmarks demonstrate that our model exhibits great generalization ability in diverse scenes and can reconstruct higher-frequency details with less memory and computational consumption.

\clearpage  

\section*{Acknowledgements}

This work is financially supported by Outstanding Talents Training Fund in Shenzhen, Shenzhen Science and Technology Program-Shenzhen Cultivation of Excellent Scientific and Technological Innovation Talents project(Grant No. RCJ-C20200714114435057), Shenzhen Science and Technology Program-Shenzhen Ho-ng Kong joint funding project (Grant No. SGDX20211123144400001), National Natural Science Foundation of China U21B2012, R24115SG MIGU-PKU META VISION TECHNOLOGY INNOVATION LAB, Guangdong Provincial Key Laboratory of Ultra High Definition Immersive Media Technology. 
Jianbo Jiao is supported by the Royal Society Short Industry Fellowship (SIF$\setminus$R1$\setminus$231009).
In addition, we sincerely thank all assigned anonymous reviewers in ECCV 2024, whose comments were constructive and very helpful to our writing and experiments.

%
%
\bibliographystyle{splncs04}
\bibliography{main}

\clearpage

\renewcommand{\thesection}{\Alph{section}}
\renewcommand{\thetable}{\Alph{table}}
\renewcommand{\thefigure}{\Alph{figure}}
\renewcommand{\theequation}{\alph{equation}}

\title{Supplementary Materials}

\titlerunning{Surface-Centric Modeling for High-Fidelity Surface Reconstruction}

\author{}

\authorrunning{R. Peng et al.}

\institute{}

\maketitle

\setcounter{section}{0}
\setcounter{figure}{0}
\setcounter{table}{0}
\setcounter{equation}{0}

\section{Results of Fine-tuning}
\label{sec:finetune}

\begin{figure*}[h]
\centering
    \includegraphics[trim={23cm 11cm 25cm 7cm},clip,width=\textwidth]{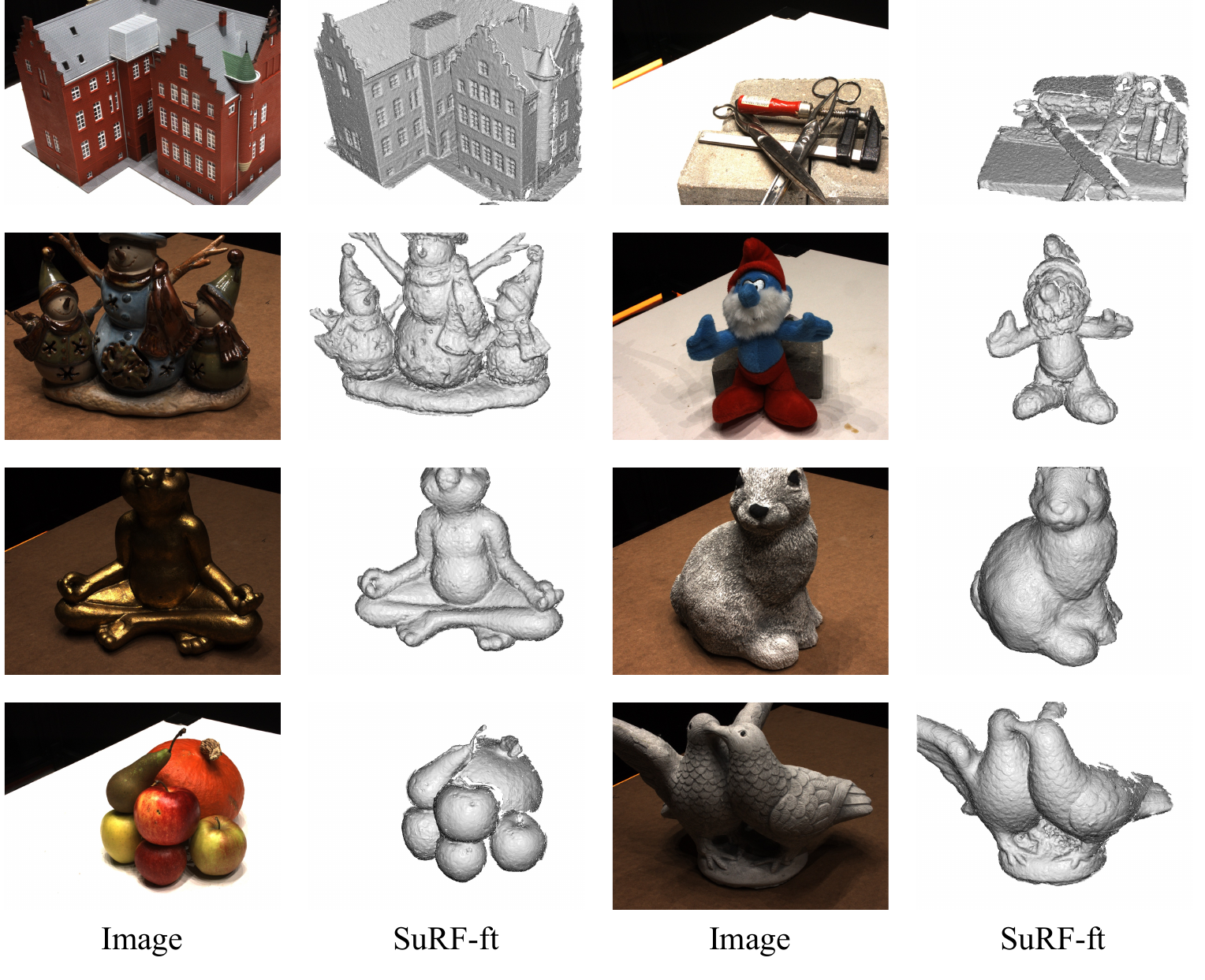}
    \vspace{-20pt}
    \caption{{\bf Visualization of fine-tuning results.}}
    \label{fig:supp_finetune}
\end{figure*}

As the qualitative and quantitative comparisons shown in our main paper, the reconstructions of our model without fine-tuning exhabit finest geometric details even compared with some fine-tuned models like SparseNeuS-ft \cite{long2022sparseneus}. Here, we illustrate some results of our model after fast fine-tuning. Different with methods \cite{xu2023c2f2neus,johari2022geonerf} that require reconstructing separate cost volume for each view, our model only builds the global volume, which makes our model easily fine-tuned (only 2.5k iterations, about 10 minutes). The quantitative results in Tab. \ref{tab:supp_finetune} show that our model still ranks the first in most scenes and has the best mean chamfer distance. Meanwhile, it is worth noting that our volume is sparse and more memory and computationally efficient. And the qualitative results of some scenes are visulized in Fig. \ref{fig:supp_finetune}. Note that there are only three input views during fine-tuning.

\begin{table}[h]
  \caption{{\bf Quantitative results of the fine-tuned model on DTU dataset.} Best results in each category are in {\bf bold} and the second best are in \underline{underline}.}
  \label{tab:supp_finetune}
  \centering
  \resizebox{1.0\linewidth}{!}{
  \begin{tabular}{l|ccccccccccccccc|c}
  \toprule
  Method & 24 & 37 & 40 & 55 & 63 & 65 & 69 & 83 & 97 & 105 & 106 & 110 & 114 & 118 & 122 & Mean \\
  \midrule
  IBRNet-ft \cite{wang2021ibrnet} & 1.67 & 2.97 & 2.26 & 1.56 & 2.52 & 2.30 & 1.50 & 2.05 & 2.02 & 1.73 & 1.66 & 1.63 & 1.17 & 1.84 & 1.61 & 1.90 \\
  SparseNeuS-ft \cite{long2022sparseneus} & 1.29 & \underline{2.27} & 1.57 & 0.88 & 1.61 & 1.86 & 1.06 & 1.27 & 1.42 & 1.07 & 0.99 & 0.87 & 0.54 & 1.15 & 1.18 & 1.27 \\
  GenS-ft \cite{peng2023gens} & \underline{0.91} & 2.33 & \underline{1.46} & \textbf{0.75} & \textbf{1.02} & \underline{1.58} & \underline{0.74} & \underline{1.16} & \underline{1.05} & \textbf{0.77} & \underline{0.88} & \underline{0.56} & \textbf{0.49} & \textbf{0.78} & \textbf{0.93} & \underline{1.03} \\
  SuRF-ft (Ours) & \textbf{0.73} & \textbf{2.11} & \textbf{1.39} & \underline{0.83} & \underline{1.05} & \textbf{1.53} & \textbf{0.68} & \textbf{1.03} & \textbf{1.02} & \underline{0.84} & \textbf{0.85} & \textbf{0.46} & \textbf{0.49} & \underline{0.84} & \underline{1.00} & \textbf{0.99} \\

  \bottomrule
  \end{tabular}
  }
  \vspace{-25pt}
\end{table}

\begin{table*}[h]
  \caption{{\bf Reproduced results of VolRecon \cite{ren2023volrecon} and ReTR \cite{liang2023rethinking} in two image sets.}}
  \label{tab:supp_res_volrecon_retr}
  \centering
  \resizebox{1.0\linewidth}{!}{
  \begin{tabular}{c|c|ccccccccccccccc|c}
  \toprule
  Method & Image Set & 24 & 37 & 40 & 55 & 63 & 65 & 69 & 83 & 97 & 105 & 106 & 110 & 114 & 118 & 122 & Mean \\
  \hline
  \multirow{3}{*}{VolRecon \cite{ren2023volrecon}} & Set0 & 1.27 & 2.66 & 1.54 & 1.04 & 1.41 & 1.94 & 1.10 & 1.53 & 1.36 & 1.08 & 1.18 & 1.37 & 0.74 & 1.22 & 1.26 & 1.38 \\
  & Set1 & 1.80 & 3.46 & 2.14 & 1.12 & 1.92 & 1.74 & 1.17 & 1.72 & 1.63 & 1.31 & 0.94 & 1.46 & 0.78 & 1.23 & 1.30 & 1.58 \\
  & Average & 1.54 & 3.05 & 1.84 & 1.08 & 1.67 & 1.84 & 1.13 & 1.63 & 1.49 & 1.19 & 1.06 & 1.42 & 0.76 & 1.22 & 1.29 & 1.48 \\
  \hline
  \multirow{3}{*}{ReTR \cite{liang2023rethinking}} & Set0 & 1.05 & 2.32 & 1.47 & 0.97 & 1.22 & 1.52 & 0.88 & 1.30 & 1.29 & 0.87 & 1.07 & 0.76 & 0.58 & 1.11 & 1.12 & 1.17 \\
  & Set1 & 1.42 & 2.95 & 1.76 & 0.99 & 1.55 & 1.59 & 0.92 & 1.49 & 1.50 & 1.19 & 0.79 & 0.89 & 0.60 & 1.09 & 1.21 & 1.33 \\
  & Average & 1.23 & 2.63 & 1.62 & 0.98 & 1.38 & 1.56 & 0.90 & 1.39 & 1.39 & 1.02 & 0.93 & 0.83 & 0.59 & 1.10 & 1.16 & 1.25 \\
  \bottomrule
  \end{tabular}
  }
  \vspace{-25pt}
\end{table*}

\section{More Comparisons with RC-MVSNet}

\begin{figure*}[t]
\centering
    \includegraphics[trim={10.5cm 15cm 5cm 13cm},clip,width=1.0\linewidth]{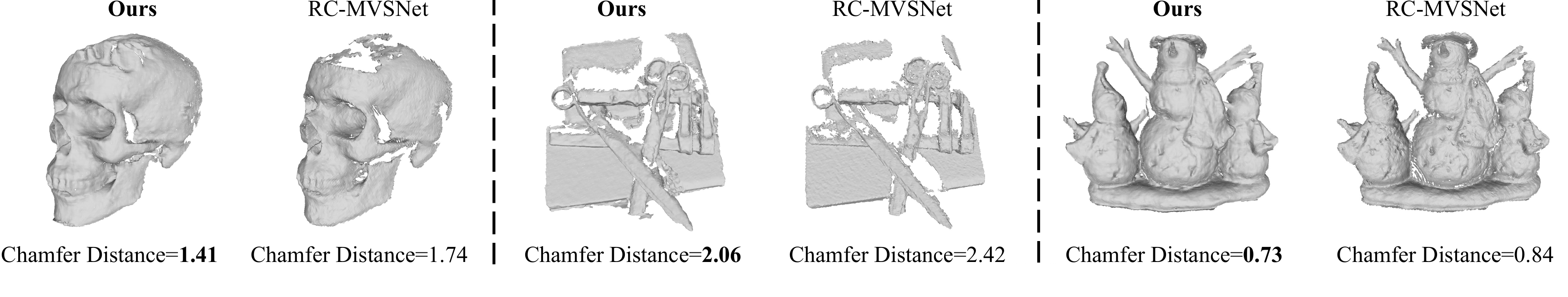}
    \vspace{-20pt}
    \caption{{\bf Visual and metric comparisons with RC-MVSNet.}}
    \vspace{-15pt}
    \label{fig:supp_comp_rcmvsnet}
\end{figure*}

We show some visual and metric comparisons with the TSDF fusion result of RC-MVSNet \cite{chang2022rc} in Fig. \ref{fig:supp_comp_rcmvsnet}. The results show that the reconstruction of our model is smoother and more complete, especially in low-texture regions, leading to better results in the chamfer distance metric. To further verify the effectiveness of our surface-centric modeling, we compare with two baselines which directly use the surface point of RC-MVSNet to prune voxels: Baseline1 directly replaces the surface region of our trained model with that of RC-MVSNet; Baseline2 uses the surface region of RC-MVSNet to train a new model. Results in Tab. \ref{tab:supp_ablawrcmvsnet} show that even simply using the surface region of RC-MVSNet can achieve superior results. And our full model, trained together with the surface location module (Our matching field), achieves the best performance. This is reasonable because the surface region of these two baselines was not optimized or corrected with the model when directly using the results of RC-MVSNet. 

\begin{table}[t]
  \caption{{\bf More ablation studies about directly using the surface region of RC-MVSNet.}}
  \vspace{-10pt}
  \label{tab:supp_ablawrcmvsnet}
  \begin{center}
  \resizebox{0.7\linewidth}{!}{
  \begin{tabular}{l|c|c|c|c}
  \toprule
   & RC-MVSNet & Baseline1 & Baseline2 & Ours \\
  \hline
  Chamfer Distance & 1.22 & 1.13 & 1.16 & \textbf{1.05} \\
  \bottomrule
  \end{tabular}
  }
  \end{center}
  \vspace{-20pt}
\end{table}

\section{Detailed Results of VolRecon and ReTR}
\label{sec:volrecon_retr}

As mentioned in our main paper, we report the reproduced results of VolRecon \cite{ren2023volrecon} and ReTR \cite{liang2023rethinking} on two image sets using their official repositories and released model checkpoints. The detailed reproduction results of all scenes at two image sets are illustrated in Tab. \ref{tab:supp_res_volrecon_retr}, which are slightly different from the results reported in their papers. We speculate that there are something inconsistent in the experimental configurations, but this inconsistency doesn't affect the valuable of their contributions.

\section{More Ablation Results}
Here, we report more ablation results of our model, and we set the training time to a quarter of the overall process (different from our main paper to save time) and only test on the first image set for convenience.

\begin{wraptable}{r}{0.5\columnwidth}
  \vspace{-33px}
  \caption{{\bf Ablation results on DTU dataset.} }
  \vspace{5px}
  \label{tab:supp_ablation}
  \centering
  \resizebox{1.0\linewidth}{!}{
  \begin{tabular}{c|c|c|c}
  \toprule
  Number of scales & Surface sampling & Cross-scale fusion & Mean \\
  \hline
  1 scales & \ding{51} & \ding{51} & 1.38 \\
  2 scales & \ding{51} & \ding{51} & 1.22 \\
  3 scales & \ding{51} & \ding{51} & 1.15 \\
  4 scales & \ding{51} & \ding{51} & \textbf{1.11} \\
  5 scales & \ding{51} & \ding{51} & 1.13 \\
  \hline
  4 scales & \ding{51} & \ding{55} & 1.15 \\
  4 scales & \ding{55} & \ding{51} & 1.13 \\
  \bottomrule
  \end{tabular}
  }
  \vspace{-15px}
\end{wraptable}

\begin{figure*}[h]
\centering
    \includegraphics[trim={8.5cm 5cm 15cm 8.5cm},clip,width=\textwidth]{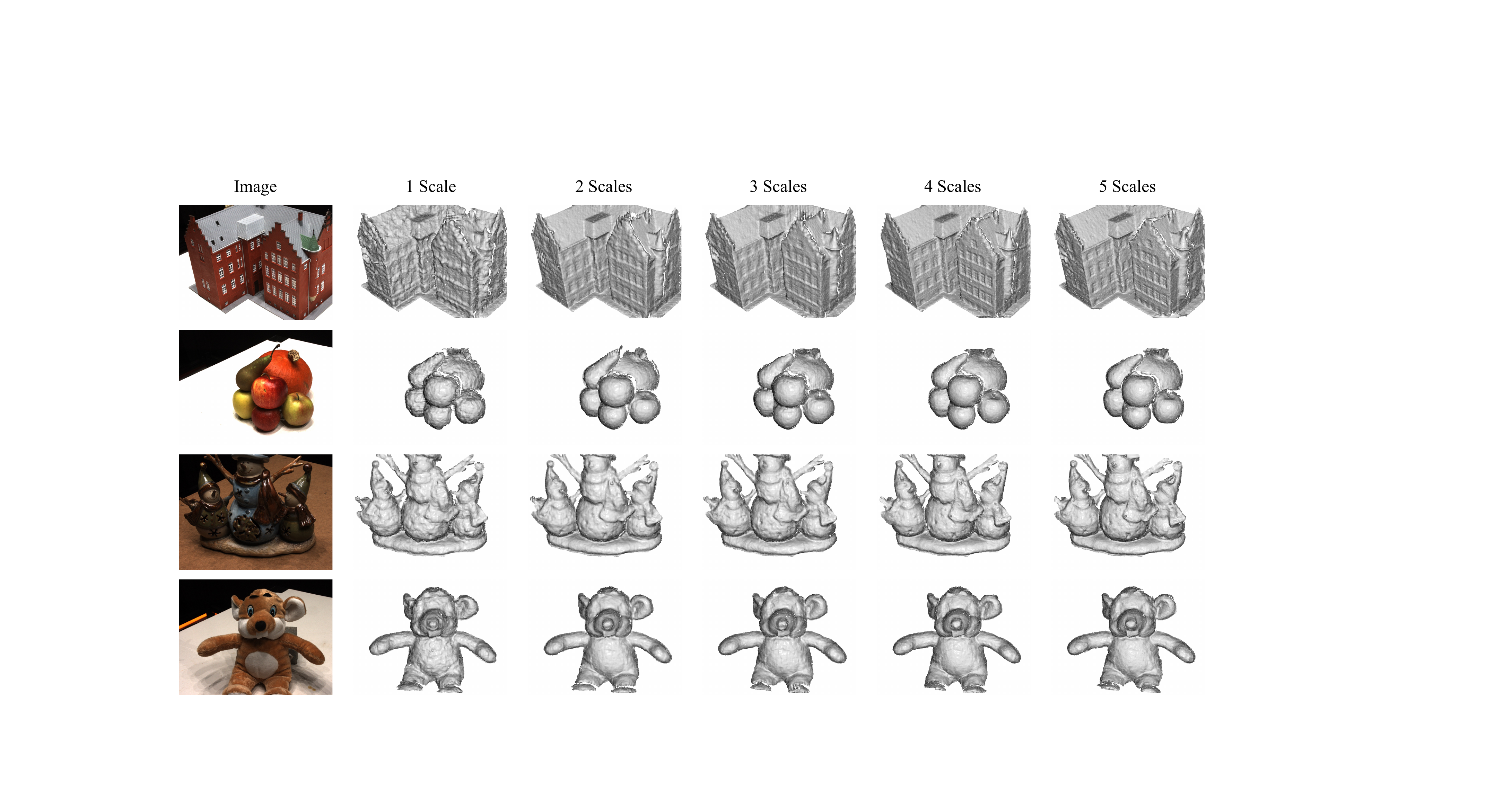}
    \vspace{-10pt}
    \caption{{\bf Visual comparison with different number of scales on DTU dataset.}}
    \vspace{-20pt}
    \label{fig:supp_num_stages}
\end{figure*}

\noindent
{\bf Number of scales.} We conduct some ablations to evaluate the effect of the number of scales. We set the resolution of the finest stage of each model to be similar. The results in Tab. \ref{tab:supp_ablation} show that the overall quality first remarkably increases and then slightly decreses, reaching the optimum in 4 scales. We illustrate some visual results of the model with differnet scales in Fig. \ref{fig:supp_num_stages}. The single-scale model performs the worst, with reconstructions that are noisy and lack geometric detail, while the four-scales model can reconstruct smooth geometry and restore more geometric details.

\begin{wraptable}{r}{0.5\columnwidth}
  \vspace{-33px}
  \caption{{\bf Ablation results of loss weight on DTU dataset.} }
  \vspace{5px}
  \label{tab:supp_loss_weight}
  \centering
  \resizebox{1.0\linewidth}{!}{
  \begin{tabular}{c|c|c|c}
  \toprule
  Method & $\beta$ & $\mu^1 : \mu^2 : \mu^3 : \mu^4$ & Mean \\
  \hline
  A & 0.0 & $0.25:0.50:0.75:1.00$ & 1.17 \\
  B & 1.0 & $0.25:0.50:0.75:1.00$ & \bf{1.11} \\
  C & 1.0 & $1.00:1.00:1.00:1.00$ & 1.16 \\
  D & 1.0 & $1.00:0.75:0.50:0.25$ & 1.25 \\
  \bottomrule
  \end{tabular}
  }
  \vspace{-15px}
\end{wraptable}

\begin{figure}[h]
\centering
    \vspace{-10pt}
    \includegraphics[trim={7cm 20cm 19cm 10cm},clip,width=\linewidth]{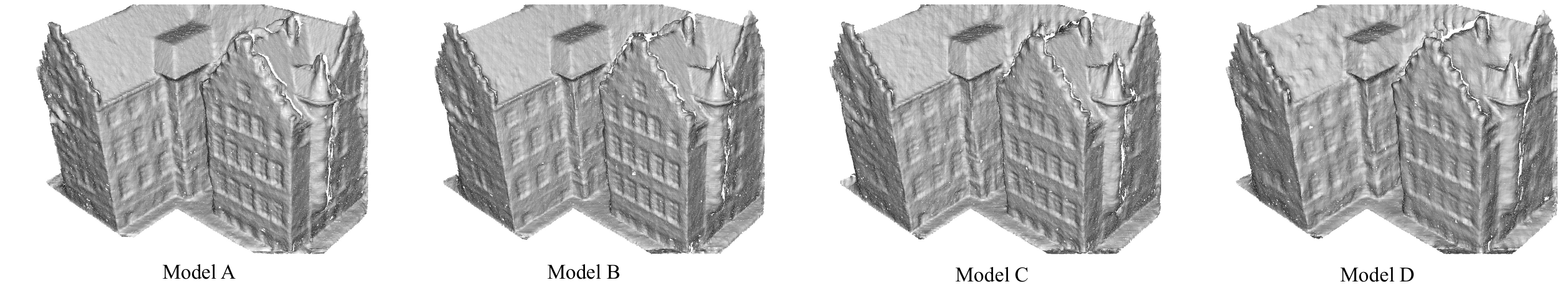}
    \vspace{-15pt}
    \caption{{\bf Visual comparison of reconstructions with different loss weights.}}
    \vspace{-20pt}
    \label{fig:supp_loss_weight}
\end{figure}

\noindent
{\bf Ablations on loss weight.} We further conduct some experiments to verify the effect of the weight of each loss term on model performance. Concretely, we change the weight of the pseudo loss $\beta$ and the weight combination of different stages of matching field loss $\mu^j$. The ablation model is based on the 4-scales model and the results are shown in Tab. \ref{tab:supp_loss_weight}. Through the comparison between model A and model B, we can see that the pseudo point clouds generated from the unsupervised multi-view stereo method \cite{chang2022rc} can guide the model towards better convergence. To avoid the influence of erroneous pseudo points, we apply a very strict filtering strategy, \ie, Only point clouds whose projection distance from at least 3 viewpoints does not exceed 0.2 pixels and whose relative depth error does not exceed 0.001 can be left. From the results of the model (B, C, D) adopt different weight combinations of the matching field loss, we can see that model B which has $\mu^1 : \mu^2 : \mu^3 : \mu^4=0.25:0.50:0.75:1.00$ performs the best and model D performs the worst. This indicates that applying greater weight to the high-resolution scale is beneficial to model convergence. Because there is no need to obtain very accurate predictions in the low-resolution scale, and the gradient of the high-resolution scale will be transmitted back to the low-resolution scale, it is reasonable to have a lower weight in the low-resolution scale. And we show some visual comparisons of these models in Fig. \ref{fig:supp_loss_weight}.

\begin{wraptable}{r}{0.5\columnwidth}
  \vspace{-33px}
  \caption{{\bf Ablation results of the range of surface regions.}}
  \vspace{5px}
  \label{tab:supp_range}
  \centering
  \resizebox{1.0\linewidth}{!}{
  \begin{tabular}{c|c|c}
  \toprule
  Method & $\epsilon^1 : \epsilon^2 : \epsilon^3 : \epsilon^4$ & Mean \\
  \hline
  B & $1.00:0.40:0.10:0.01$ & 1.11 \\
  E & $1.00:0.30:0.10:0.01$ & \bf{1.10} \\
  F & $1.00:0.30:0.05:0.01$ & 1.12 \\
  \bottomrule
  \end{tabular}
  }
  \vspace{-20px}
\end{wraptable}

\noindent
{\bf Ablations on the range of surface region.} Here, we employ an additional ablation experiment to study the sensitivity of the range of surface regions. $\epsilon^0$ is the range of the first scale, and its value is fixed at 1, which represents covering the entire near and far area. And the value of later scales means the percentage of coverage. As the results shown in Tab. \ref{tab:supp_range}, the differences of these three groups of experiments are not large, as long as the surface region is gradually tightened, and model F which has a range combination of $\epsilon^1 : \epsilon^2 : \epsilon^3 : \epsilon^4=1.00:0.30:0.10:0.01$ performs the best. 

\section{More Results}
\label{sec:more_resuts}

\begin{figure*}[h]
\centering
    \vspace{-20pt}
    \includegraphics[trim={21cm 4.5cm 24.3cm 1.5cm},clip,width=\textwidth]{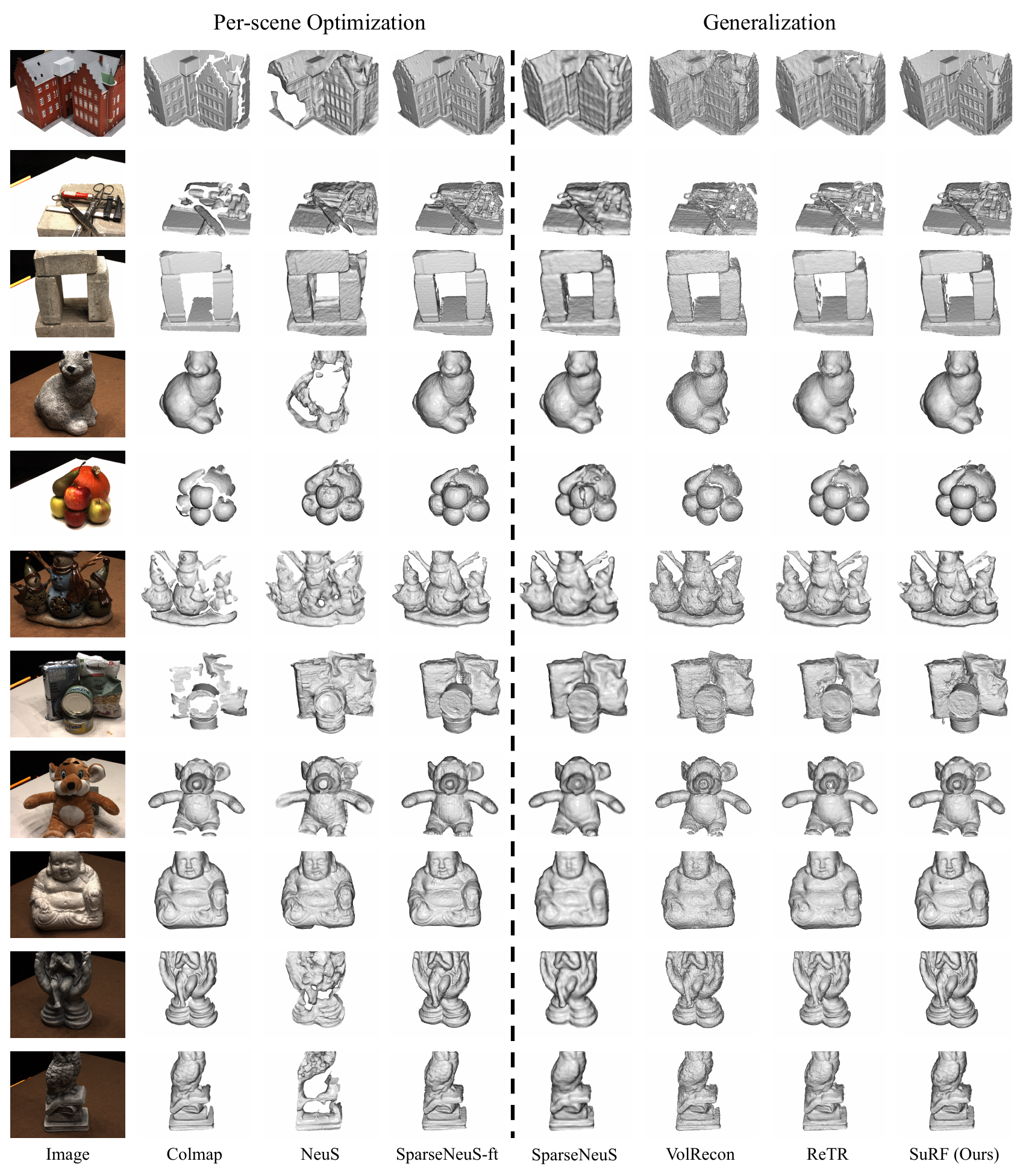}
    \vspace{-15pt}
    \caption{{\bf More qualitative comparisons on DTU dataset.}}
    \vspace{-15pt}
    \label{fig:supp_more_results}
\end{figure*}

\begin{figure*}[t]
\centering
    \includegraphics[trim={1cm 19cm 25cm 12cm},clip,width=\textwidth]{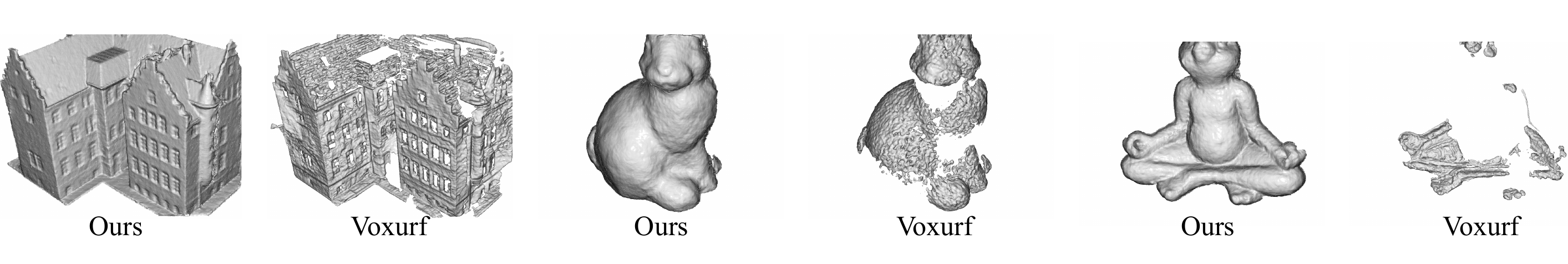}
    \vspace{-15pt}
    \caption{{\bf Comparison with Voxurf on DTU dataset with 3 inputs.}}
    \label{fig:supp_comp_voxurf}
\end{figure*}

Because C2F2NeuS \cite{xu2023c2f2neus} doesn't release the code, the memory of C2F2NeuS in Tab. \ref{tab:mem_view} is refer to the implementation of CasMVSNet \cite{gu2020cascade}. Fig. \ref{fig:supp_more_results} shows additional comparisons with COLMAP \cite{schonberger2016structure}, NeuS \cite{wang2021neus}, SparseNeuS \cite{long2022sparseneus}, SparseNeuS-ft \cite{long2022sparseneus}, VolRecon \cite{ren2023volrecon} and ReTR \cite{liang2023rethinking} on DTU dataset. We can see that our method can stably achieve superior results and exhibit finer geometry details. We further show some visual comparisons with the fast per-scene overfitting method Voxurf \cite{wu2022voxurf} in Fig. \ref{fig:supp_comp_voxurf}. While Voxurf still requires more than 30 minutes of training time per scene, it struggles to reconstruct smooth and accurate surface from sparse inputs. 

\section{Visualization of the Surface Region}
\label{sec:vis_surf}

\begin{figure*}[t]
\centering
    \includegraphics[trim={3.5cm 13.5cm 1.5cm 12cm},clip,width=\textwidth]{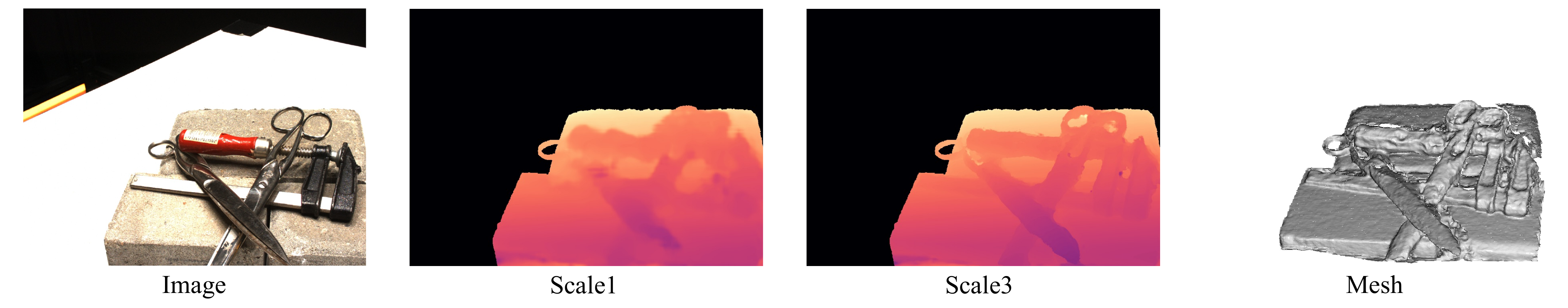}
    \vspace{-15pt}
    \caption{{\bf Visualization of the located surface region from the matching field at different scales.} We visualize the depth of the middle surface for convenience. 
    }
    \vspace{-10pt}
    \label{fig:supp_surface_region}
\end{figure*}

To understand how the surface region changes as scale increases, we show some visualization results of the surface region at different scales in Fig. \ref{fig:supp_surface_region}. For convenience, we show the depth of the middle position of the surface region. We can see that the located surface region at the higher resolution scale is indeed sharper, which proves the effectiveness of our design. 

\section{Limitations and Future Work}
\label{sec:limitation}

Despite exhibiting efficiency over existing methods, our model still struggled to extract the surface in real time due to the inherent drawback of MLP-based implicit methods. In the future, we will be focusing on addressing this deficiency issue, and we have constructed a lite-version model, which will be released latter. Furthermore, we plane to train our model on more large-scale dataset like Objaverse \cite{deitke2023objaverse} and expand the scale of the model like \cite{hong2023lrm,li2023instant3d}.

\end{document}